\documentclass[journal]{IEEEtran}

\ifCLASSINFOpdf
\else
\fi

\usepackage{color,soul}
\usepackage{graphicx,amsmath,amsfonts}
\usepackage{tkz-tab}
\usepackage{multirow}
\usepackage{subcaption}
\usepackage{steinmetz}

\usepackage{listings}

\definecolor{dkgreen}{rgb}{0,0.6,0}
\definecolor{gray}{rgb}{0.5,0.5,0.5}
\definecolor{mauve}{rgb}{0.58,0,0.82}

\begin{document}

\title{Do Neural Network Weights account for Classes Centers?}

\author{Ioannis~Kansizoglou,
        Loukas~Bampis,
        and~Antonios~Gasteratos,~\IEEEmembership{Senior~Member,~IEEE}
\thanks{The authors are with the Department of Production and Management Engineering, Democritus University of Thrace, Xanthi 67100, Greece e-mail: (see https://robotics.pme.duth.gr).}
}
\markboth{Journal of \LaTeX\ Class Files,~Vol.~XX, No.~XX, April~2021}%
{Shell \MakeLowercase{\textit{et al.}}: Bare Demo of IEEEtran.cls for IEEE Journals}
\maketitle

\begin{abstract}
The exploitation of Deep Neural Networks (DNNs) as descriptors in feature learning challenges enjoys apparent popularity over the past few years.
The above tendency focuses on the development of effective loss functions that ensure both high feature discrimination among different classes, as well as low geodesic distance between the feature vectors of a given class.
The vast majority of the contemporary works rely their formulation on an empirical assumption about the feature space of a network's last hidden layer, claiming that \textit{the weight vector of a class accounts for its geometrical center in the studied space}.
The paper at hand follows a theoretical approach and indicates that the aforementioned hypothesis is not exclusively met.
This fact raises stability issues regarding the training procedure of a DNN, as shown in our experimental study.
Consequently, a specific symmetry is proposed and studied both analytically and empirically that satisfies the above assumption, addressing the established convergence issues.
\end{abstract}


\IEEEpeerreviewmaketitle

\section{Introduction}\label{Intro}

\IEEEPARstart{E}{fficient} data representation into a compact metric space occupies prominent place in the recent challenges of Computer Science (CS)~\cite{bengio2013representation,zhong2016overview}.
To that end, effort is concentrated on mapping each pattern of the sensory input into a space displaying specific and known metric properties through a suitably designed method.
Such a method may derive either from traditional feature engineering or from a learning-based algorithm~\cite{ma2020image, liang2017text}.
In the first case, the output representation is usually obtained via a mathematical projection rule, like principal component analysis~\cite{pearson1901liii} and linear discriminant analysis~\cite{fisher1936use}, or a specific extraction scheme making use of a set of pre-defined handcrafted rules~\cite{lowe2004distinctive, bay2006surf, eyben2010opensmile,tsintotas2019probabilistic}.
In contrast, the second case employs a machine learning algorithm, in order to discover salient features from raw data~\cite{norouzi2011minimal, khalid2014survey, latif2020deep}.

Due to the complex nature of real-world sensory inputs in present problems, the manual definition of descriptive features becomes increasingly unreliable, leaving no option but to engage machine learning schemes.
Typical algorithms are Support Vector Machines (SVMs)~\cite{norouzi2011minimal} and Deep Neural Networks (DNNs)~\cite{latif2020deep, guo2019deep}, with DNNs forming the leading choice in feature extraction for cascade~\cite{caicedo2015active,balaska2021enhancing} and fusion tasks~\cite{antol2015vqa,kansizoglou2019active} given their proven efficacy over the past years.
Yet, in order for a learning algorithm to ensure robust representation capacity, a set of techniques needs to be applied during its training phase.
The above techniques form the main research subject of \textit{feature} or \textit{representation learning}~\cite{bengio2013representation}.
The studied assessment criteria for a method are mainly focused on its capability to map similar patterns into close geodesic distance in the output metric space, while keeping high distance between unlike inputs.
The above properties are broadly known as high \textit{intra-class compactness} and \textit{inter-class discrepancy}~\cite{deng2019arcface}, respectively, where class refers to the set of all the database instances from a common label category.
Hence, the mutual satisfaction of the above two criteria is incorporated into the optimization goal of the adopted learning algorithm~\cite{deng2019arcface, liu2017sphereface}.

The most prevailing approaches in the field of feature learning have been developed in image retrieval problems and more specifically in the challenge of face verification~\cite{masi2018deep}.
The adopted Convolutional Neural Network (CNN) architectures are trained with an enhanced version of the original Softmax loss, which embodies the two representation learning criteria, by shaping a space with specific geometrical properties.
Given a training step, each feature vector, formed by the activations of the last hidden layer, is forced to approximate the centre of its target class.
Based on this rule, an ample variety of approaches have been developed outweighing prior works in the field~\cite{deng2019arcface, liu2017sphereface, wang2018cosface}.
However, in order to materialize a representation for each class into a computationally effective loss function, the weight vector of each class is utilized, assuming that it coincides with the respective center.
The above practice shapes a hypothesis which is from now on referred as $\mathcal{H}$. 

The present study sheds light on the empirical supposition $\mathcal{H}$ since it is not exclusively met for arbitrary distributions of the weight vectors, introducing an implicit error in the training procedure.
Moving further, a specific symmetry regarding those vectors in the studied feature space is proposed, in order to satisfy $\mathcal{H}$.
Eventually, we proceed with empirical findings about the impact of the added error during the training procedure of a DNN.

The remainder of this paper is structured as follows.
In Section II, we discuss typical approaches in the field of discriminative feature learning that denote the assumption $\mathcal{H}$, while Section III clearly states the motivation behind the current work. 
Then, Sections IV and V theoretically prove the inconsistency between a class's weight vector and its geometrical center for an arbitrary distribution of the weights, as well as propose a novel symmetrical layout.
Section VI provides a detailed description regarding the implementation of the above symmetry.
Subsequently, Section VII displays several experiments, proving the efficiency of the proposed layer and illustrating the way that the implicit error of the initial supposition affects the DNN's convergence.
In the last section, the acquired conclusions are collected, suggesting concise concepts that can be employed in representation learning tasks.

\section{Related Work} \label{RelWor}

In this section, a comprehensive review is presented, focusing on the exploitation of DNNs in feature learning challenges.
Consequently, individual attention is paid to recent trends in verification tasks, as well as the theoretical background that forms the basis of the proposed methods.

\subsection{Neural-based Feature Learning} \label{RelWorkA}

The original implementation of the Softmax loss has been progressively proved insufficient for challenging verification tasks that require high feature discrimination.
Motivated by this fact, a host of methods have been developed, proposing the enhancement of the common loss with advanced restrictive conditions.
The initial step, introduced with the concept of \textit{Siamese Networks}, suggests the exploitation of two or more identical configurations of a DNN during the training procedure~\cite{koch2015siamese}.
Hence, the extracted feature vectors can be compared, by feeding two or more samples, respectively, passing their similarity score to a suitable loss function under a pairwise learning procedure.
In such a procedure, the common cross-entropy loss can not be implemented, leading to the utilization of novel objective functions, with \textit{triplet}~\cite{schroff2015facenet} and \textit{contrastive loss}~\cite{hadsell2006dimensionality} forming the most typical ones.
At each training step, the objective of \textit{triplet loss} refers to: i) a distance minimization component between a specific training sample and another one of the same class (\textit{positive pair}), as well as ii) a distance maximization component between the same sample and a sample of another class (\textit{negative pair})~\cite{hoffer2015deep}.
Aiming at an improved inter-class separation, the \textit{contrastive loss} compares for each training sample the similarity score of a positive pair against a negative one~\cite{hadsell2006dimensionality}.

\subsection{Enhanced-Softmax Loss Functions} \label{RelWorkB}

Despite their efficiency, Siamese Networks displayed two main disadvantages, \textit{viz.}, increased training duration, due to pairwise learning, and opaque outputs since they do not provide a probabilistic distribution, like the Softmax loss.
Thus, the research community has resorted to alternative solutions that exploit Softmax loss and enhance its discrimination capacity.
Accordingly, the \textit{center loss} proposed the penalization of a feature vector by means of its Euclidean distance from the center of its target class~\cite{wen2016discriminative}.
During training, the center of each class was calculated and updated by the respective feature vectors of the mini-batch, adding a considerable computational cost in the procedure.
Two variations of \textit{center loss}, \textit{viz.}, the \textit{island} and \textit{range loss}, were developed in order to handle the above issues~\cite{cai2018island, zhang2017range}.

In an effort to reduce the required computational complexity, the idea of applying more rigorous constraints directly in the Softmax loss, rather than providing complementary losses, was introduced.
On this account, Large margin Softmax (\textit{L-Softmax}) encouraged both discrimination criteria in the common cross-entropy loss, by adjusting an angular margin that scales the angle between a feature and the weight vector of its target class~\cite{liu2016large}.
Thereby, the weight vector comprises the reference point of a class, or equally, the desired orientation for the feature vectors.
Instead of an angular constraint, \mbox{\textit{L$_2$-Softmax}} suggested the feature vectors' normalization, in order for them to lie on the surface of a hypersphere with configurable radius, boosting the discrimination performance~\cite{wang2017normface}.
Combining both angular and norm constraints, \textit{SphereFace} proposed an improved version of the original Softmax loss, \textit{aka.} \mbox{\textit{A-Softmax}}~\cite{liu2017sphereface}.
\textit{SphereFace} was also the first method that coped with stability issues, requiring the supervision of a simple Softmax loss mainly during the initial steps of the training procedure.
\textit{CosFace} inserted an additive cosine margin directly to the target logits of the CNN as an attempt to improve stability~\cite{wang2018cosface}.
Subsequently, \textit{ArcFace} succeeded an even simpler and simultaneously efficient approach that adopts an additive angular margin~\cite{deng2019arcface}.
However, all of the above methods keep the same strategy of applying an angular margin between the feature and the weight vector of the target class, assuming that this is its optimal orientation~\cite{deng2019arcface, liu2017sphereface, wang2018cosface}.
An entirely different approach constitutes the implementation of the angular margin astride the decision boundaries between the classes~\cite{kansizoglou2020haseparator}.

\section{Motivation}

\begin{figure}
    \centering
    \begin{subfigure}[b]{0.24\textwidth}
        \centering
        \includegraphics[width=0.9\linewidth]{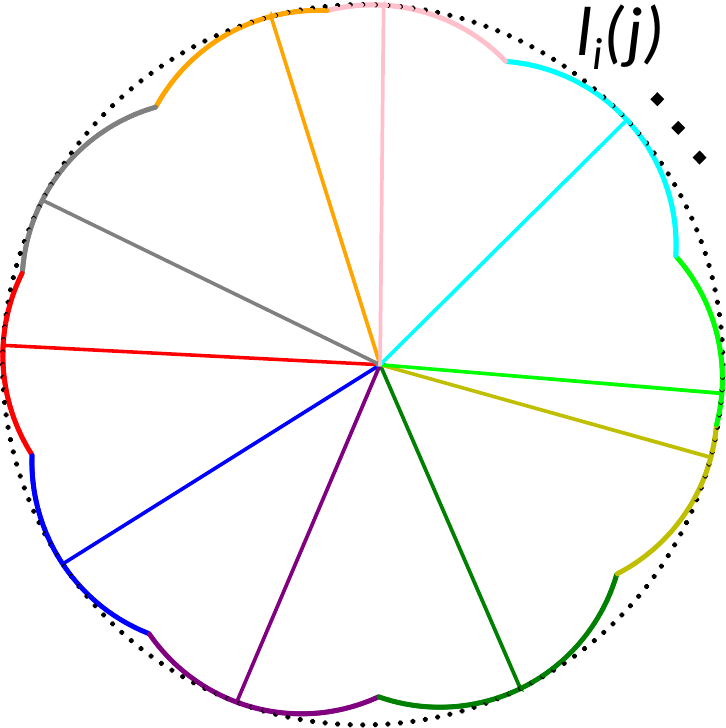}
        \caption{Dot product outputs}
        \label{fig:MotiveA}
    \end{subfigure}%
	~
    \begin{subfigure}[b]{0.24\textwidth}
        \centering
        \includegraphics[width=0.9\linewidth]{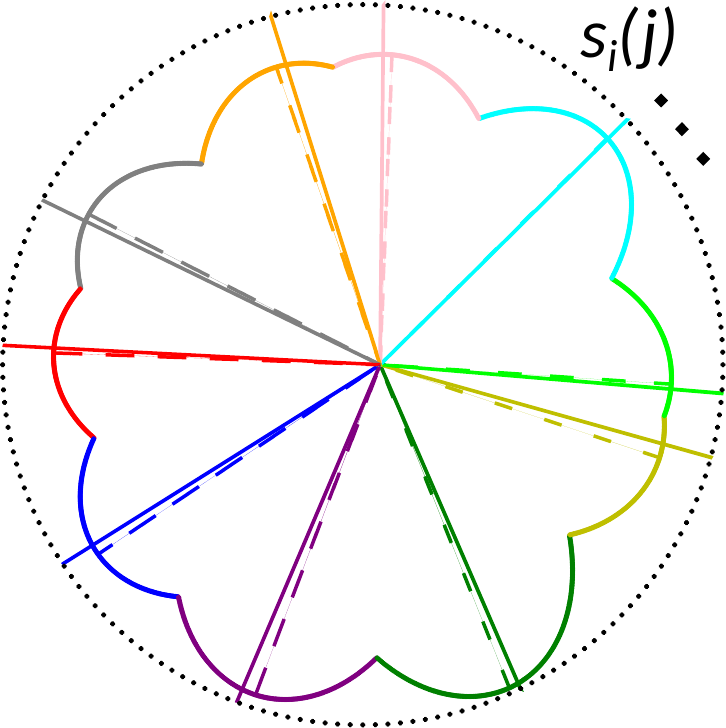}
        \caption{Softmax outputs}
        \label{fig:MotiveB}
    \end{subfigure}
    \caption{\small (a) The resulting dot product distribution of $l_i(j)$ and its extreme points (continuous lines) training a vanilla CNN with L$_2$-constrained Softmax on MNIST~\cite{lecun1998gradient}.
(b) The resulting softmax distribution of $s_i(j)$ for the same network. 
Note that, the extreme points of softmax outputs (dashed lines) diverge from the corresponding dot product ones (continuous lines).}
    \label{fig:Motive}
\end{figure}

Keeping in mind the stated hypothesis $\mathcal{H}$, we first present the inspiration behind the current work and formulate our established concern.
For visualization purposes, we proceed with a typical analysis in a 2-D feature space $\mathcal{F}_2\subset\mathbb{R}^2$, training a vanilla CNN on the MNIST database~\cite{lecun1998gradient} using a two-units hidden layer before the output one.
The produced weight vectors' layout is assessed in Fig.~\ref{fig:Motive}.

More specifically, given the resulting weight vectors \mbox{$\hat{w}_i\in\mathcal{F}_2$}, $i=1,2,...,10$ and the set of possible embeddings $\hat{e}(j)=1 \phase{j\frac{\pi}{180}}$, $j\in\mathbb{N}^{<360}$ with $\hat{e}(j)\in\mathcal{F}_2$ in polar coordinates, we calculate the dot product output values:
\begin{equation}
l_i(j) = \max_i{(\hat{w}_i\cdot\hat{e}(j))}.
\end{equation}
Also $\forall j$ we keep the index $i$ that corresponds to the maximum value, denoting the prevailing class.
As proved in a previous work of ours~\cite{kansizoglou2020deep}, each class ends up occupying a convex angular subspace in $\mathcal{F}_2$, while the extreme points of $l_i(j)$ calculated by:
\begin{equation}
dl_i(j) / dj = 0
\end{equation}
coincide with the orientation of the weight vectors $\hat{w}_i$, as shown in Fig.~\ref{fig:MotiveA}.
By further applying the common Softmax activation function and keeping its maximum value, defined as:
\begin{equation}
s_i(j) = \max_i \left(\frac{e^{\hat{w}_i\cdot\hat{e}_j}}{\sum_k^{10}{e^{\hat{w}_k\cdot\hat{e}_j}}}\right),
\end{equation}
we produce the regarding softmax distribution, depicted in Fig.~\ref{fig:MotiveB}.
By calculating the orientations $j$ that account for the extreme points of $s_i(j)$ from:
\begin{equation}
ds_i(j) / dj = 0,
\end{equation}
we find out that it diverges from the equivalent dot product ones.
Bearing that in mind, the reader can understand the error introduced by the use of $\hat{w}_i$ as the point of reference in a loss function, especially considering a space $\mathcal{F}$ of increased dimensionality, as well as the randomness of $\hat{w}_i$ during the initialization of a training procedure.

\section{Rules in High Dimensional Spaces}

We proceed with several findings that enable the study of an arbitrary feature space $\mathcal{F}_d\subset\mathbb{R}^{d+1}$, $d\in\mathbb{N}^{>1}$.

\textit{\textbf{Lemma 1}:}
\textit{Given the vectors $\bar{v}_1,\bar{v}_2,...,\bar{v}_n$, $n\in\mathbb{N}^{>1}$ with equal norms in the 2-D plane $P$, such that $\widehat{(\bar{v}_i,\bar{v}_{i+1})}=2\pi/n$, \mbox{$\forall i=1,...,n-1$}, then:}
\begin{equation}\label{eq:Lemma1}
\sum_{i=1}^{n}{\bar{v}_i}=\bar{0}.
\end{equation}
\textit{Proof}:
The above is trivial to be shown through the proof by contradiction.
Let us suppose the vectors $\bar{v}_i=\overline{OV}_i$ of equal norm in $P$, with $i=1,2,...,n$ and their sum $\sum_{i=1}^{n}{\bar{v}_i}=\bar{v}$, where $\bar{v}$ also lies in $P$.
If we rotate the plane $P$ -- or equally all vectors $\bar{v}_i$ -- with respect to their common origin $O$ by an angle of $2\pi /n$, then their layout does not change, implying that their sum should also remain unchanged.
However, $\bar{v}$ is also rotated by $2\pi/n$, indicating that the initial sum changes, except for the case that $\bar{v}=\bar{0}$.

\textit{\textbf{Lemma 2}:}
\textit{Given the equation:
\begin{equation}\label{eq:Lemma2}
\sum_{k=0}^{n-1}{\sin(x-2k\pi/n)e^{\cos(x-2k\pi/n)}}=0,\textit{ } n\in\mathbb{N}^{>1},
\end{equation}
$x_r = \frac{2r\pi}{n}$, $r\in\mathbb{N}^{<n}$ are roots of Eq. \ref{eq:Lemma2} in $[0,2\pi)$.}\\
\textit{Proof}:
Without loss of generality, let us consider the root $x_{r_0}=2r_0\pi/n$, with $r_0\in\mathbb{N}^{(1,n/2)}$.
It is trivial to show that the term for $k=r_0$ in Eq.~\ref{eq:Lemma2} is equal to zero.
Then, Eq.~\ref{eq:Lemma2} becomes:
\begin{multline}
\sum_{k=0}^{n-1}{\sin(2r_0\pi/n-2k\pi/n)e^{\cos(2r_0\pi/n-2k\pi/n)}}=0 \implies \\
\implies \sum_{k=0}^{r_0-1}{\sin(2(r_0-k)\pi/n)e^{\cos(2(r_0-k)\pi/n)}} \\
+\sum_{k=r_0+1}^{n-1}{\sin(2(r_0-k)\pi/n)e^{\cos(2(r_0-k)\pi/n)}}=0.
\end{multline}
For the values of $k$ astride $r_0$: $k^{\pm}_{m_0}=r_0\pm m_0$, $\forall m_0\in\mathbb{N}^{(0,r_0]}$, we have:
\begin{multline}
\sin(2(r_0-(r_0-m_0))\pi/n)e^{\cos(r_0-(r_0-m_0))\pi/n)} \\
+\sin(2(r_0-(r_0+m_0))\pi/n)e^{\cos(r_0-(r_0+m_0))\pi/n)} = \\
\\= \sin(2m_0\pi/n)e^{\cos(2m_0\pi/n)} + \sin(-2m_0\pi/n)e^{\cos(-2m_0\pi/n)} = \\
\\= \sin(2m_0\pi/n)e^{\cos(2m_0\pi/n)} - \sin(2m_0\pi/n)e^{\cos(2m_0\pi/n)} = 0.
\end{multline}
Ergo, the astride $2r_0$ terms plus the $k=r_0$ itself, \textit{i.e.}, $2r_0+1$ terms of Eq.~\ref{eq:Lemma2}, are eradicated, leading to the remaining summation terms:
\begin{equation}
\sum_{k=2r_0+1}^{n-1}{\sin(2(r_0-k)\pi/n)e^{\cos(2(r_0-k)\pi/n)}}=0.
\end{equation}
We further examine the terms $\mathcal{I}^{+}_{m_1}$ and $\mathcal{I}^{-}_{m_1}$ for $k^{+}_{m_1}=2r_0+1+m_1$ and $k^{-}_{m_1}=n-1-m_1$, respectively, $\forall m_1\in\mathbb{N}^{<(n-2r_0-1)/2}$.
\begin{multline}
\mathcal{I}^{+}_{m_1}: \sin(2(r_0-2r_0-1-m_1)\pi/n)e^{\cos(2(r_0-2r_0-1-m_1)\pi/n)} = \\
= \sin(-2(r_0+m_1+1)\pi/n)e^{\cos(-2(r_0+m_1+1)\pi/n)} = \\
= -\sin(2(r_0+m_1+1)\pi/n)e^{\cos(2(r_0+m_1+1)\pi/n)}.
\end{multline}
\begin{multline}
\mathcal{I}^{-}_{m_1}: \sin(2(r_0-n+1+m_1)\pi/n)e^{\cos(2(r_0-n+1+m_1)\pi/n)} = \\
= \sin(2(r_0+m_1+1)\pi/n-2\pi)e^{\cos(2(r_0+m_1+1)\pi/n-2\pi)} = \\
= \sin(2(r_0+m_1+1)\pi/n)e^{\cos(2(r_0+m_1+1)\pi/n)}.
\end{multline}
Hence, $\mathcal{I}^{+}_{m_1}+\mathcal{I}^{-}_{m_1}=0$, $\forall m_1\in\mathbb{N}^{<(n-2r_0-1)/2}$.
We discern two cases for even and odd values of $n-2r_0-1$, respectively.
In the first case, the total terms of Eq.~\ref{eq:Lemma2} have been eradicated and the lemma holds.
In case that $n-2r_0-1$ is odd, then a singular term remains that is the center of $[0,n-2r_0-1)$, or equally, for $m=(n-2r_0-2)/2=-(r_0+1)+n/2$.
Note that:
\begin{equation}
\begin{split}
k^{+}_m: 2r_0+1+(-r_0-1+n/2)=r_0+n/2\\
k^{-}_m: n-1-(-r_0-1+n/2)=r_0+n/2.
\end{split}
\end{equation}
However, for this term we have:
\begin{multline}
\sin(2(r_0-(r_0+n/2))\pi/n)e^{\cos(2(r_0-(r_0+n/2))\pi/n)} = \\
 = \sin(-\pi)e^{\cos(-\pi)} = 0,
\end{multline}
concluding that our statement holds for the second case, as well.
Finally, following the similar procedure it is trivial to show that the same conclusion holds for $n/2<r_0<n$.

\begin{figure}
    \centering
    \includegraphics[width=0.85\linewidth]{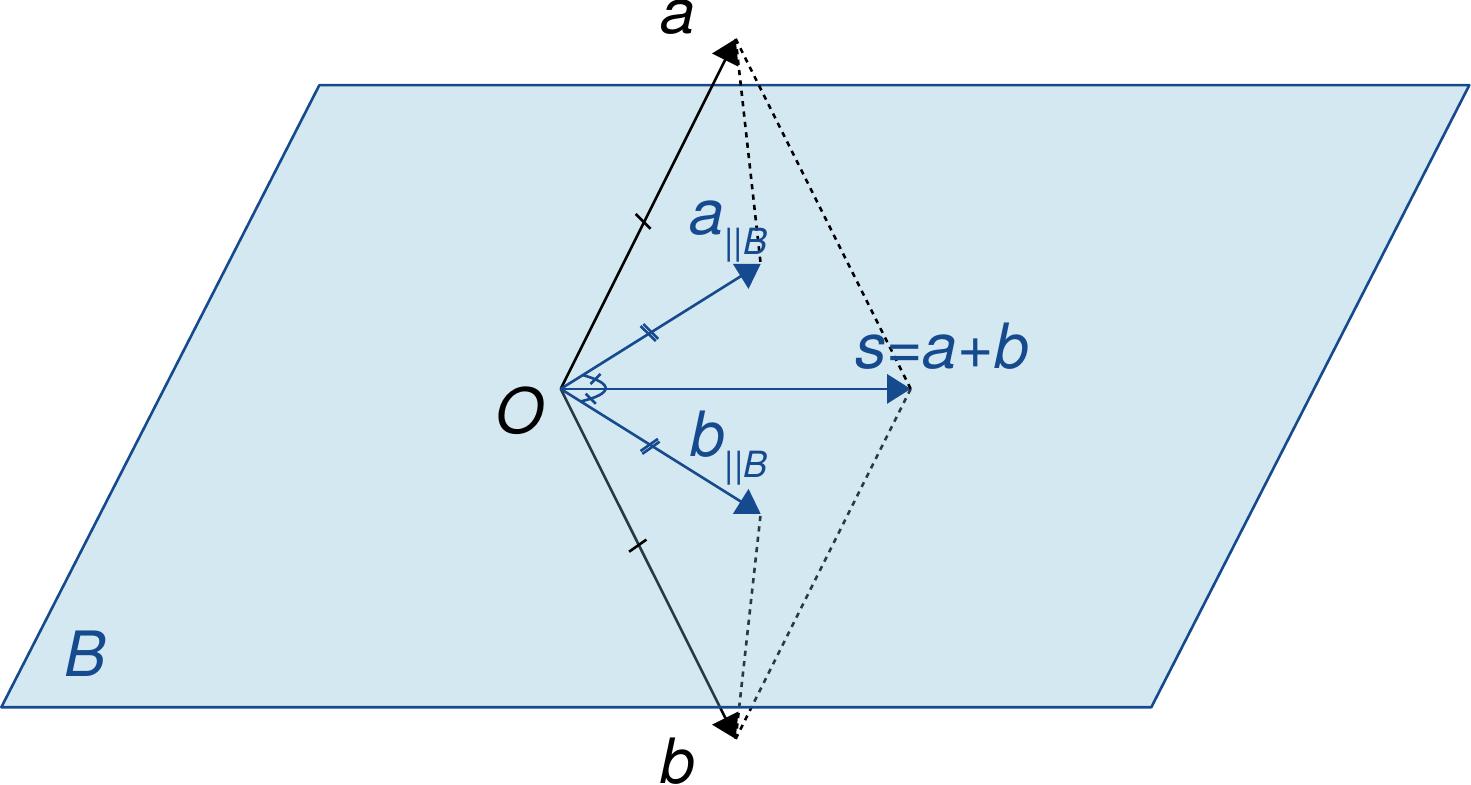}
    \caption{\small Geometrical interpretation of \textit{\textbf{Lemma 3}}.}
    \label{fig:Lemma3}
\end{figure}

\textit{\textbf{Lemma 3}}: \textit{Let two unit vectors $a$, $b\in\mathcal{F}_d$ with a common origin $O$, as well as a bivector $B$ that passes through their summation vector $s=a+b$, then the projections of $a$, $b$ onto $B$ have equal norms and angular distances from $s$.}\\
\textit{Proof (based on rules of Clifford Algebra~\cite{clifford1882classification})}: From~\cite{suter2003geometric}, $a_{\parallel_B} = (a\cdot B)B^{-1}$ and $b_{\parallel_B} = (b\cdot B)B^{-1}$.
Given that $B$ passes through $s$, we know that:
\begin{equation}
sB = s\cdot B+ s\wedge B = s\cdot B.
\end{equation}
For the addition of the collinear with $B$ vectors, we have:
\begin{multline}\label{lem1}
a_{\parallel_B}+b_{\parallel_B} = (a\cdot B+b\cdot B)B^{-1} = \\
= ((a +b)\cdot B)B^{-1} = (s\cdot B)B^{-1} = \\
= (sB)B^{-1} = s.
\end{multline}
Moreover for the projections of $a, b$ on $s$, we write:
\begin{gather*}
\left.
\begin{array}{ll}
a_{\parallel_s}=a\cdot s = a\cdot(a+b) = a\cdot a + a\cdot b = 1 + a\cdot b \\
b_{\parallel_s}=b\cdot s = b\cdot(a+b) = b\cdot a + b\cdot b = 1 + a\cdot b \\
\end{array} 
\right\} \implies
\end{gather*}
\begin{gather}
\implies a_{\parallel_s} = a\cdot s = b\cdot s = b_{\parallel_s}.
\end{gather}
Hence,
\begin{gather*}
a\cdot s = b\cdot s \implies\\
\implies(a_{\parallel_B}+a_{\perp_B})\cdot s = (b_{\parallel_B}+b_{\perp_B})\cdot s\implies\\
\implies a_{\parallel_B}\cdot s+a_{\perp_B}\cdot s = b_{\parallel_B}\cdot s+b_{\perp_B}\cdot s\implies\\
\implies a_{\parallel_B}\cdot s = b_{\parallel_B}\cdot s \implies
\end{gather*}
\begin{gather}
\implies(a_{\parallel_B}-b_{\parallel_B})\cdot s=0,
\end{gather}
suggesting perpendicularity between $a_{\parallel_B}-b_{\parallel_B}$ and $s$.
However, those vectors constitute the diagonals of a parallelogram $\mathcal{P}$ on $B$ with sides $a_{\parallel_B}$ and $b_{\parallel_B}$.
The above perpendicularity between the diagonals, drops $\mathcal{P}$ to the degenerate case of a \textbf{rhombus} with sides $a_{\parallel_B}$ and $b_{\parallel_B}$, thus ensuring:
\begin{itemize}
\item $\widehat{(a_{\parallel_B},b_{\parallel_B})}$ bisection by $s$, and
\item $\|a_{\parallel_B}\|=\|b_{\parallel_B}\|$.
\end{itemize}
An illustration of \textit{Lemma 3} is provided in Fig.~\ref{fig:Lemma3}.

\section{$\mathcal{H}$ as a non-exclusive condition}

In this section, we begin by demonstrating the falsifiability of $\mathcal{H}$ in the input space of the output layer $\mathcal{F}_d\subset\mathbb{R}^{d+1}$, where $d\in\mathbb{N}^{>1}$ the dimensionality both of the embeddings and the weight vectors.
Then, we proceed with the definition of a specific symmetrical layout in $\mathcal{F}_d$ that produces proven consistency with $\mathcal{H}$.
In our following analysis, we consider an arbitrary number of target classes $n\in\mathbb{N}^{>2}$.

The studied criterion, which leads to the satisfaction of $\mathcal{H}$, refers to the maximization of the $i$-th class's softmax output:
\begin{equation}
S_i = \frac{e^{z_i}}{\sum_{j=0}^{n-1}{e^{z_j}}},
\end{equation}
with $z_j = \bar{w}_j\cdot\bar{e}$, $\bar{e}\in\mathcal{F}_d$ the embedding and $\bar{w}_j\in\mathcal{F}_d$ the $j$-th class's weight vector.
Then, the criterion demands the maximization of $S_i$ for the case that $\bar{e}$ coincides with the target class's weight vector $\bar{w}_i$.
According to~\cite{kansizoglou2020deep}, the output of the $j$-th neuron can be written as $z_j = \|\bar{w}_{j_\parallel}\|\cos(\theta-\phi_j)$, where $\bar{w}_{j_\parallel}$ denotes the projection of $\bar{w}_j$ on the common plane of $\bar{w}_i$ and $\bar{e}$.
Then:
\begin{multline}
\label{eq:sym1}
\frac{dS_i}{d\theta}=0\implies\frac{d}{d\theta}\left( \frac{e^{z_i(\theta)}}{\sum_{j=1}^{n}{e^{z_j(\theta)}}}\right)=0 \implies \\
\implies \frac{de^{z_i(\theta)}}{d\theta}\sum_{j=0}^{n-1}{e^{z_j(\theta)}}-e^{z_i(\theta)}\sum_{j=0}^{n-1}{\frac{de^{z_j(\theta)}}{d\theta}}=0 \implies \\
\implies \sum_{j=0}^{n-1}{\frac{dz_i(\theta)}{d\theta}e^{z_i(\theta)}e^{z_j(\theta)}}-\sum_{j=0}^{n-1}{\frac{dz_j(\theta)}{d\theta}e^{z_i(\theta)}e^{z_j(\theta)}}=0\implies\\
\implies \sum_{j=0}^{n-1}{\left[\left(\frac{dz_i(\theta)}{d\theta}-\frac{dz_j(\theta)}{d\theta}\right)e^{z_i(\theta)}e^{z_j(\theta)}\right]} = 0.
\end{multline}
We want $S_i$ to maximize for $\bar{e}=\bar{w}_{i}$, where $\theta=\phi_i$ and $dz_i(\theta)/d\theta|_{\theta=\phi_i}=-\|\bar{w}_i\|\sin(\theta-\phi_i)|_{\theta=\phi_i} = 0$. Hence, Eq.~\ref{eq:sym1} becomes:
\begin{multline}\label{eq:sym2}
-\sum_{j=0}^{n-1}{\frac{dz_j(\theta)}{d\theta}e^{z_i(\theta)}e^{z_j(\theta)}} = 0 \implies \sum_{j=0}^{n-1}{\frac{dz_j(\theta)}{d\theta}e^{z_j(\theta)}} = 0.
\end{multline}
Eq.~\ref{eq:sym2} shapes the mathematical formulation of the studied criterion.

\subsection{Refutability of $\mathcal{H}$}

At this stage, it is sufficient to determine a layout of the weight vectors, that leads to the falsification of $\mathcal{H}$.
Let us consider $n$ weight vectors $\bar{w}_i\in\mathcal{F}_d$, \textit{s.t.}:
\begin{itemize}
\item \textit{A.I}: all lie in a 2-D plane $P$,

\item \textit{A.II}: $\|\bar{w}_i\|=1$, $\forall i=0,1,...,n-1$ and

\item \textit{A.III}: $\widehat{(\bar{w}_i,\bar{w}_{i+1})}=\pi/n$, $\forall i=0,1,...,n-2$.
\end{itemize}
By working on the common plane $P$ of \textit{A.I}, as well as taking into account \textit{A.II}, we ensure that $\|\bar{w}_{j_\parallel}\|=\|\bar{w}_j\|=1$, $\forall j\in \mathbb{N}^{<n}$ in Eq.~\ref{eq:sym2}.
Moreover, according to \textit{A.III} and by considering the reference vector $\bar{w}_0$, we have $\widehat{(\bar{w}_0,\bar{w}_{j})}=j\pi/n$, $\forall j\in \mathbb{N}^{[1,n)}$.
Elaborating more, we can express the dot product of each weight vector $\bar{w}_j$, as a function of the weight of reference $\bar{w}_0$, as follows:
\begin{gather*}
z_0 = \bar{w}_0\cdot\bar{e} = \cos\theta, \\
z_j = \bar{w}_j\cdot\bar{e} = \cos(\theta-j\pi/n),\textit{   } \forall j\in \mathbb{N}^{[1,n)}.
\end{gather*}
Hence, Eq.~\ref{eq:sym2} ends up to:
\begin{equation}\label{eq:sym3a}
\sum_{j=0}^{n-1}{\sin(\theta-j\pi/n)e^{\cos(\theta-j\pi/n)}} = 0.
\end{equation}
However, in case that $\bar{e}$ and $\bar{w}_0$ coincide, indicating $\theta=0$, we have:
\begin{equation}
\sum_{j=0}^{n-1}{\sin(j\pi/n)e^{\cos(j\pi/n)}} > 0, \forall n\in\mathbb{N}^{>2}.
\end{equation}
The above leads us to the conclusion that $\mathcal{H}$ is not satisfied given an arbitrary distribution of the weight vectors in $\mathcal{F}_d$.

\subsection{Proposed symmetry in $\mathcal{F}_d$} \label{sub:PropSym}

\begin{figure}
    \centering
    \begin{subfigure}[b]{0.24\textwidth}
        \centering
        \includegraphics[width=0.85\linewidth]{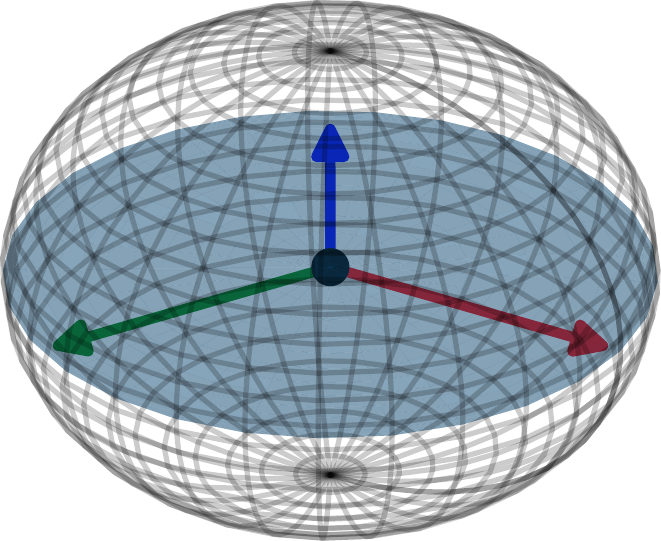}
        \caption{$n=3$}
        \label{fig:PropSymA}
    \end{subfigure}%
	~
    \begin{subfigure}[b]{0.24\textwidth}
        \centering
        \includegraphics[width=0.85\linewidth]{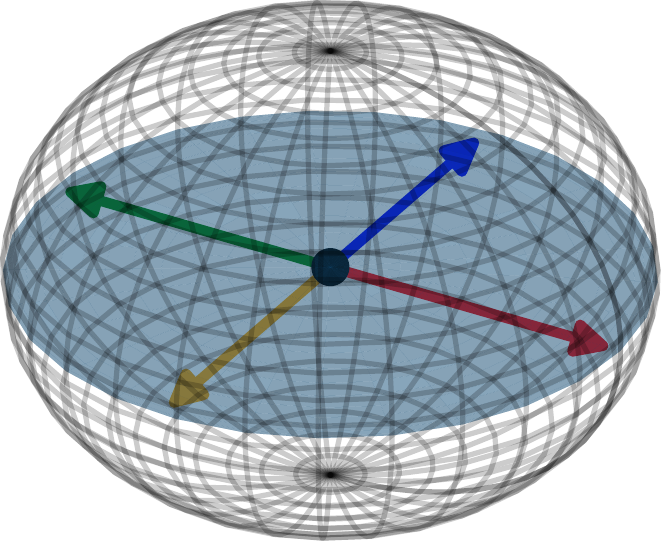}
        \caption{$n=4$}
        \label{fig:PropSymB}
    \end{subfigure}
    \caption{\small Illustration of the proposed symmetrical layout in $\mathcal{F}_3\subseteq\mathbb{R}^3$ for two different numbers of classes: (a) $n=3$ and (b) $n=4$.}
    \label{fig:PropSym}
\end{figure}

In turn, we define a specific symmetrical layout of the last layer's weights in $\mathcal{F}_d$, which ensures that those weights account for the classes' centers for any number of target classes $n$ and feature vectors' dimension $d$. 
More specifically, we examine the case of $n$ weight vectors $\bar{w}_i$ such that:
\begin{itemize}
\item \textit{B.I}: all lie in a 2-D plane $P$,

\item \textit{B.II}: $\|\bar{w}_i\|=1$, $\forall i\in\mathbb{N}^{<n}$ and

\item \textit{B.III}: $\widehat{(\bar{w}_i,\bar{w}_{i+1})}=2\pi/n$, $\forall i\in\mathbb{N}^{<n-1}$.
\end{itemize}
In Fig.~\ref{fig:PropSym}, the proposed symmetry is displayed in $\mathcal{F}_3\subseteq\mathbb{R}^3$ for (a) $n=3$ and (b) $n=4$. 

Considering the above properties, we examine the required condition to relate class centrality with its weight vector.
Given that $\bar{e}=\bar{w}_{i}$, there is no specific plane defined by $\bar{e}$ and $\bar{w}_{i}$.
Ergo, we are free to work on the common plane $P$ of \textit{B.I}, ensuring consistency with \textit{B.II} since $\|\bar{w}_{j_\parallel}\|=\|\bar{w}_j\|=1$, $\forall j\in\mathbb{N}^{<n}$ in Eq.~\ref{eq:sym2}.
Moreover, elaborating \textit{B.III}, we can express the dot product of each weight vector $\bar{w}_j$, as a function of the weight of reference $\bar{w}_0$, as follows:
\begin{gather*}
z_0 = \bar{w}_0\cdot\bar{e} = \cos\theta, \\
z_j = \bar{w}_j\cdot\bar{e} = \cos(\theta-2j\pi/n),\textit{   } \forall j\in\mathbb{N}^{[1,n)}.
\end{gather*}
Consequently, Eq.~\ref{eq:sym2} ends up to the equation:
\begin{equation}\label{eq:sym3}
\sum_{j=0}^{n-1}{\sin(\theta-2j\pi/n)e^{\cos(\theta-2j\pi/n)}} = 0.
\end{equation}
According to \textit{Lemma 2}, the weight vectors $\bar{w}_j$ satisfy Eq.~\ref{eq:sym3}, thus indicating that they constitute extremum points of $S_i$, $\forall i\in\mathbb{N}^{<n}$.
Following \textit{Lemma 1}, we can further conclude that:
\begin{itemize}
\item \textit{B.IV}: $\sum_{i=0}^{n-1}{\bar{w}_i} = \bar{0}$.
\end{itemize}

At this stage, we check the validity of Eq.~\ref{eq:sym2} on a random plane $P'$ that passes through the weight vector $\bar{w}_i$.
Let $\bar{w}_{i-m}$ and $\bar{w}_{i+m}$, $m\in\mathbb{N}^{[1,(n-1)/2)}$,  be the weight vectors astride $\bar{w}_i$  that lie on $P$. Then, $\widehat{(\bar{w}_i,\bar{w}_{i-m})}=-2m\pi/n$, $\widehat{(\bar{w}_i,\bar{w}_{i+m})}=2m\pi/n$ and:
\begin{multline}\label{eq:sym4}
\frac{dz_{i-m}(\theta)}{d\theta}e^{z_{i-m}(\theta)} + \frac{dz_{i+m}(\theta)}{d\theta}e^{z_{i+m}(\theta)} = \\
= \|\bar{w}_{i-m_\perp}\|\sin(\theta-\phi_{i-m})e^{\cos(\theta-\phi_{i-m})} \\
+\|\bar{w}_{i+m_\perp}\|\sin(\theta-\phi_{i+m})e^{\cos(\theta-\phi_{i+m})}.\\
\end{multline}
Moreover, the summation vector $\bar{w}_{i-m}+\bar{w}_{i+m}$ is parallel to $\bar{w}_i$.
According to \textit{Lemma 3} the projections of $\bar{w}_{i-m}$ and $\bar{w}_{i+m}$ on $P'$ have equal norms and are also bisected by $\bar{w}_i$.
Hence, $\|\bar{w}_{i-m_\perp}\|=\|\bar{w}_{i+m_\perp}\|=\|\bar{w}_m\|$ and $\phi_i-\phi_{i-m}=-(\phi_i-\phi_{i+m})=\phi_m$.
For $\theta=\phi_i$, Eq.~\ref{eq:sym4} becomes:
\begin{multline}
\|\bar{w}_m\|\sin(\phi_m)e^{\cos(\phi_m)} + \|\bar{w}_m\|\sin(-\phi_m)e^{\cos(-\phi_m)} = \\
= \|\bar{w}_m\|\sin(\phi_m)e^{\cos(\phi_m)} - \|\bar{w}_m\|\sin(\phi_m)e^{\cos(\phi_m)}=0.
\end{multline}
That is, the vectors that present equal angle astride $\bar{w}_i$ eradicate themselves in the sum of Eq.~\ref{eq:sym2}.
In case that their total number is odd, the remaining one is the counterbalancing vector $-\bar{w}_i$, a term which also equals zero, as shown in \mbox{\textit{Lemma 2}}.
Eventually, the proposed symmetry secures maximization of $S_i$ at the orientation of $\bar{w}_i$ from any possible direction.

\section{Implementation Details} \label{ImpDet}

\begin{figure*}
    \centering
	\begin{subfigure}[b]{0.245\textwidth}   
        \centering
        \includegraphics[width=0.95\textwidth]{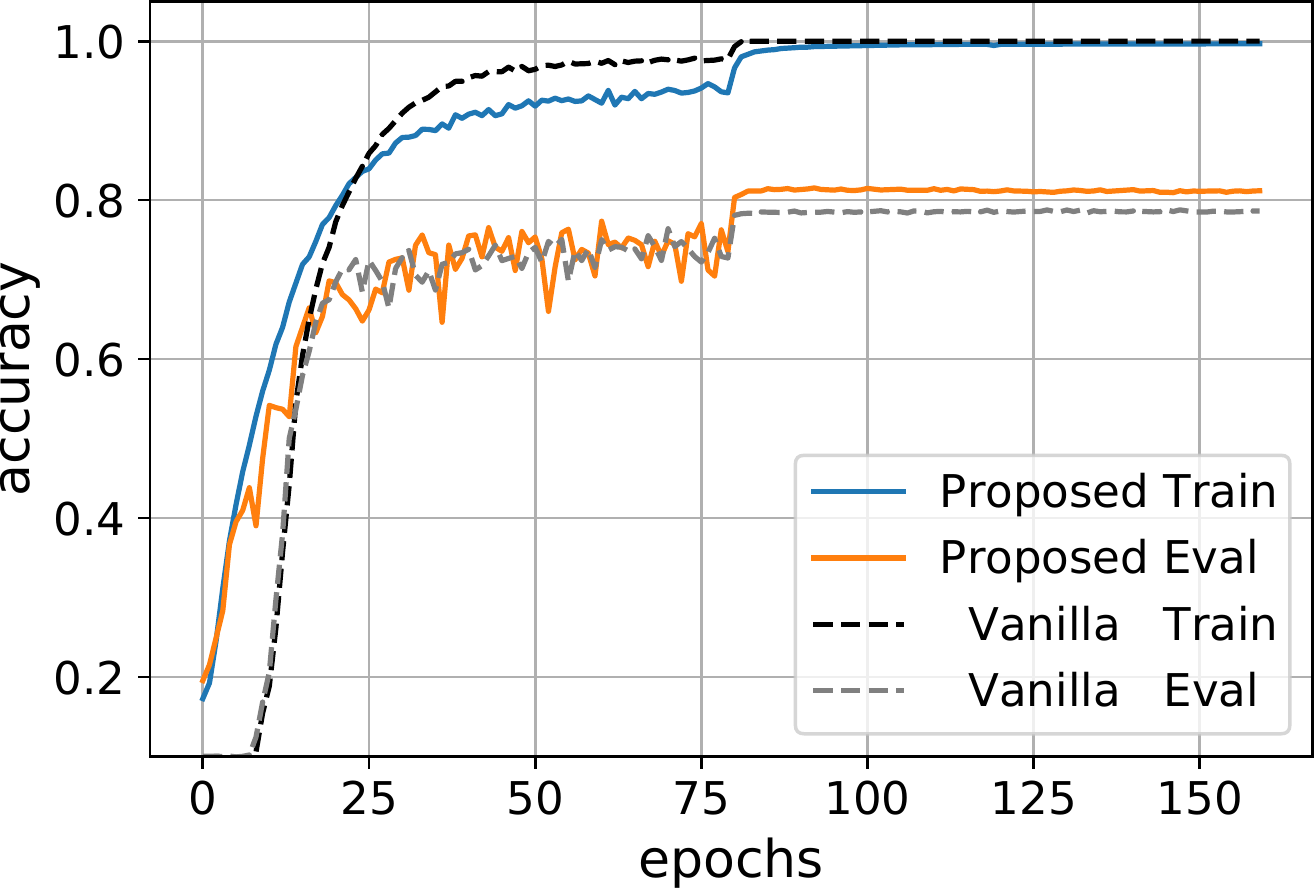}
        \caption[]{Accuracy ($\sigma$=$8$)}    
    	\label{fig:Impl_Syma}
    \end{subfigure}
    \begin{subfigure}[b]{0.245\textwidth}
        \centering
        \includegraphics[width=0.95\textwidth]{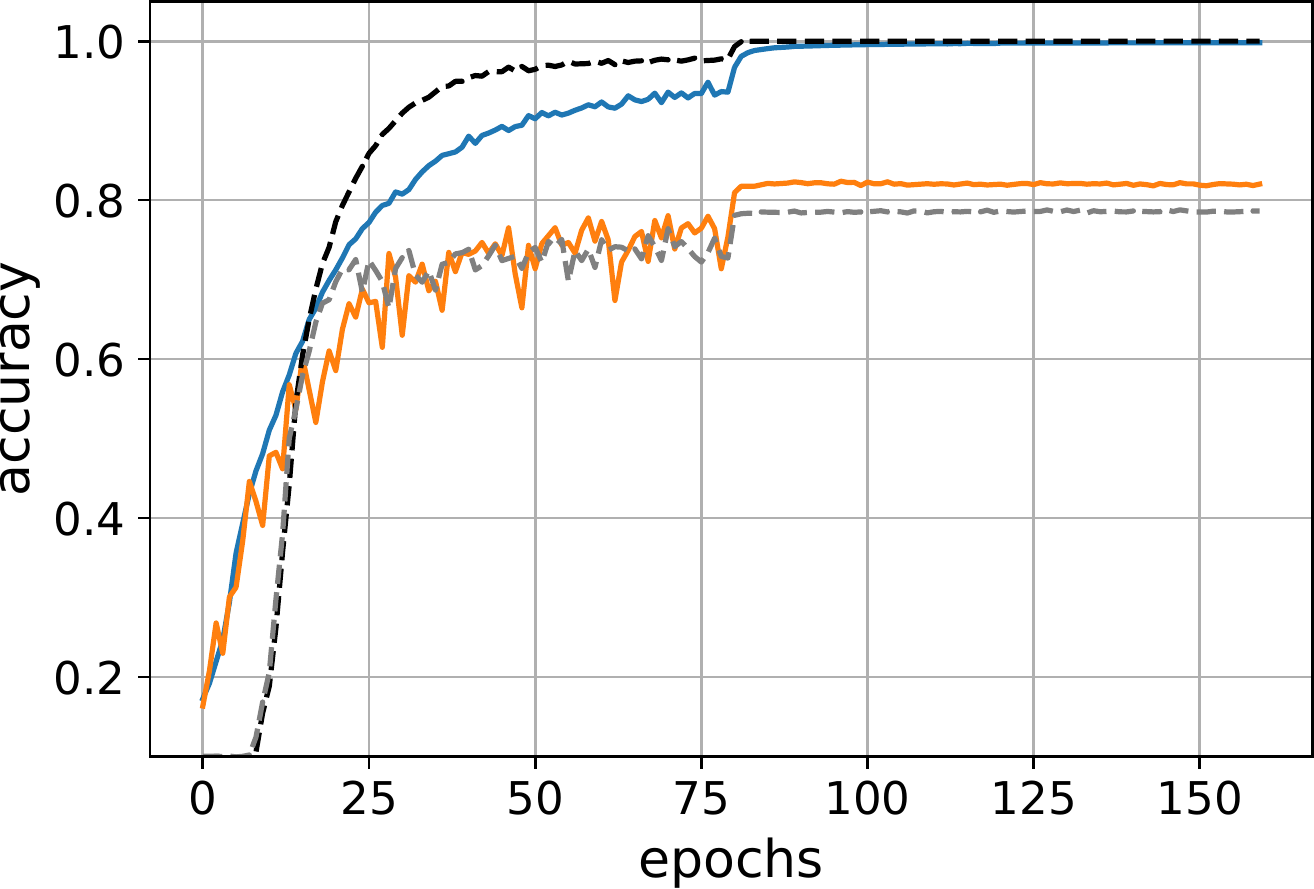}
        \caption[]{Accuracy ($\sigma$=$16$)}    
        \label{fig:Impl_Symb}
    \end{subfigure}
    \begin{subfigure}[b]{0.245\textwidth}
        \centering 
        \includegraphics[width=0.95\textwidth]{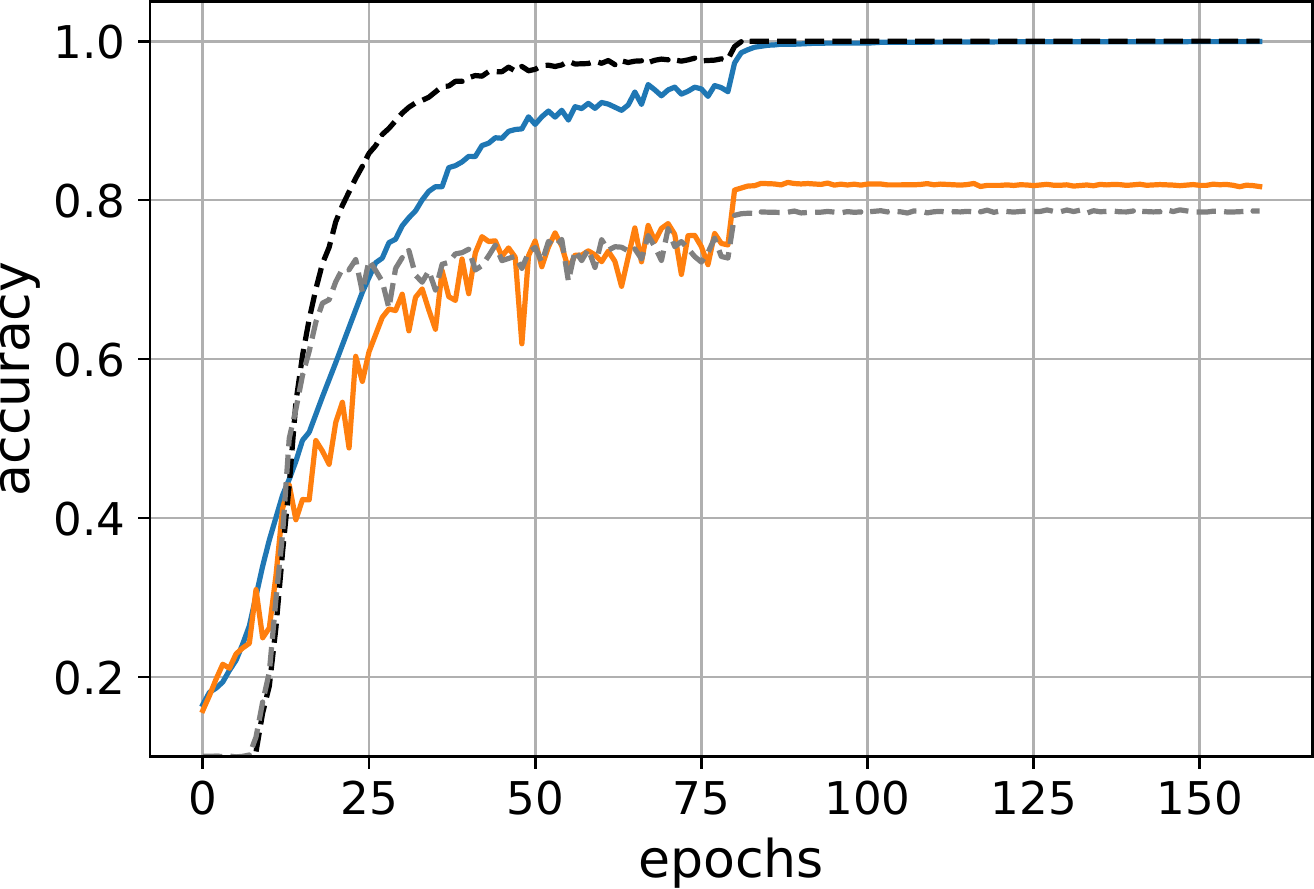}
        \caption[]{Accuracy ($\sigma$=$32$)}    
        \label{fig:Impl_Symc}
    \end{subfigure}
    \begin{subfigure}[b]{0.245\textwidth}   
        \centering 
        \includegraphics[width=0.95\textwidth]{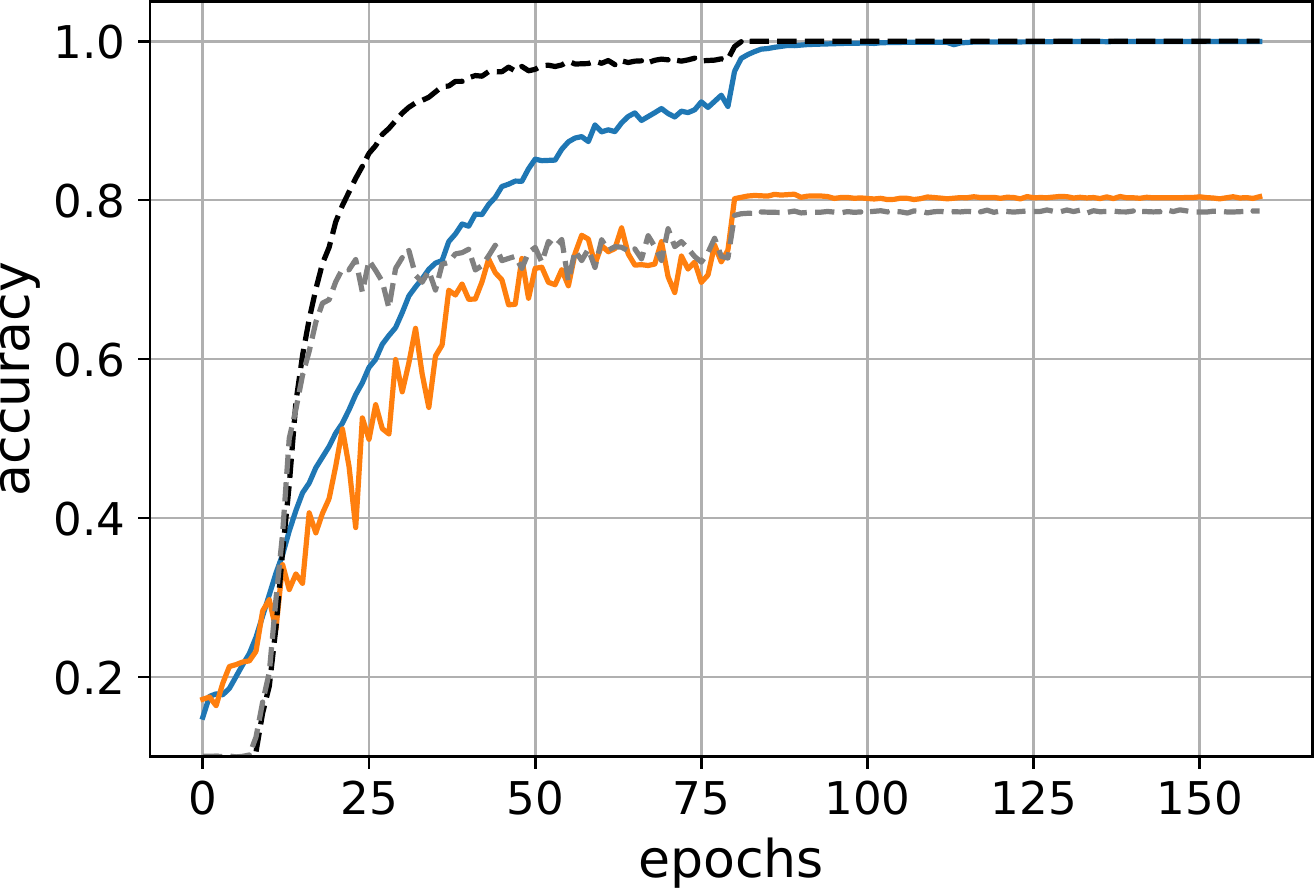}
        \caption[]{Accuracy ($\sigma$=$64$)}    
        \label{fig:Impl_Symd}
    \end{subfigure}
	\vskip\baselineskip
    \centering
	\begin{subfigure}[b]{0.245\textwidth}   
        \centering 
        \includegraphics[width=0.95\textwidth]{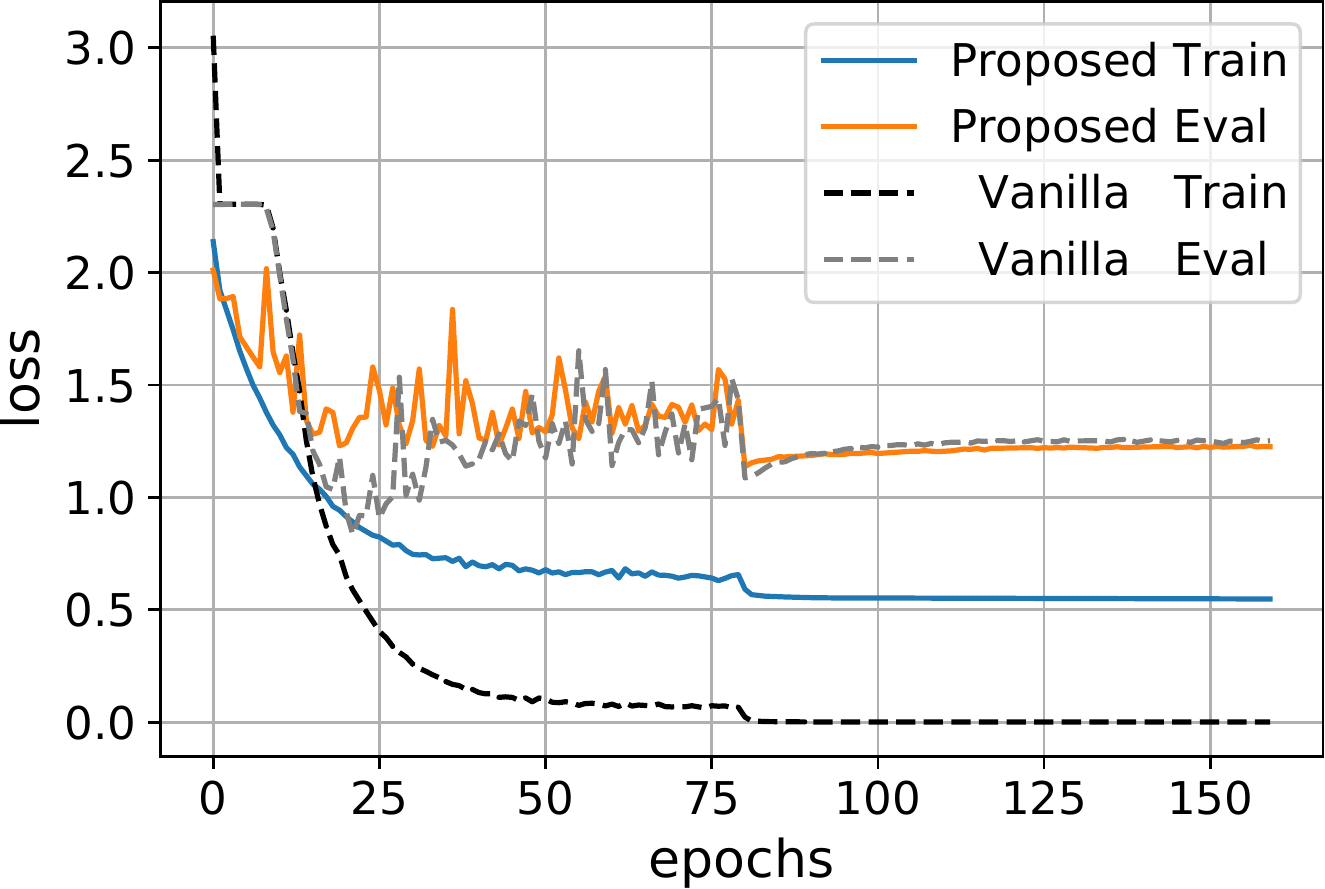}
        \caption[]{Loss ($\sigma$=$8$)}    
    	\label{fig:Impl_Syme}
    \end{subfigure}    
    \begin{subfigure}[b]{0.245\textwidth}
        \centering
        \includegraphics[width=0.95\textwidth]{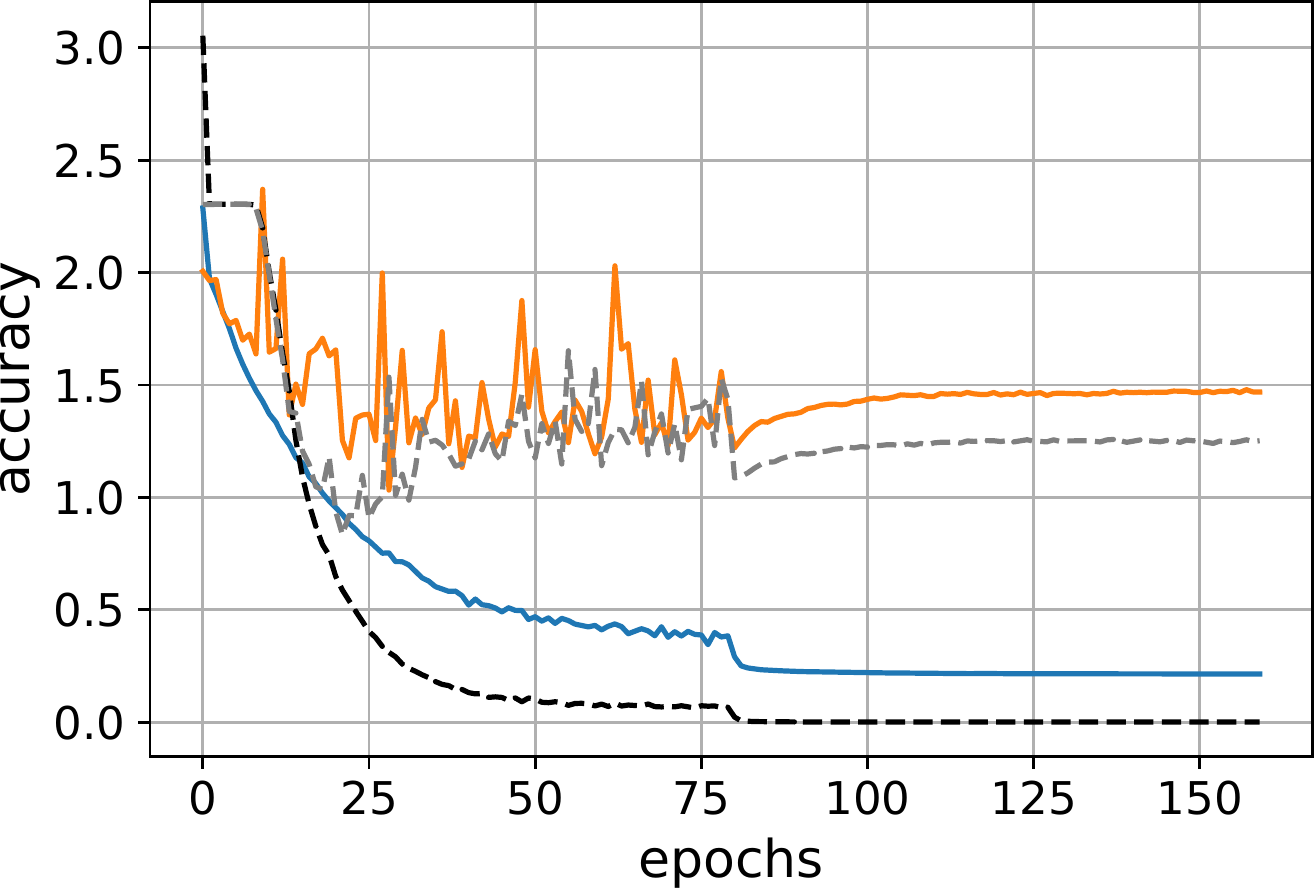}
        \caption[]{Loss ($\sigma$=$16$)}    
        \label{fig:Impl_Symf}
    \end{subfigure}
    \begin{subfigure}[b]{0.245\textwidth}
        \centering 
        \includegraphics[width=0.95\textwidth]{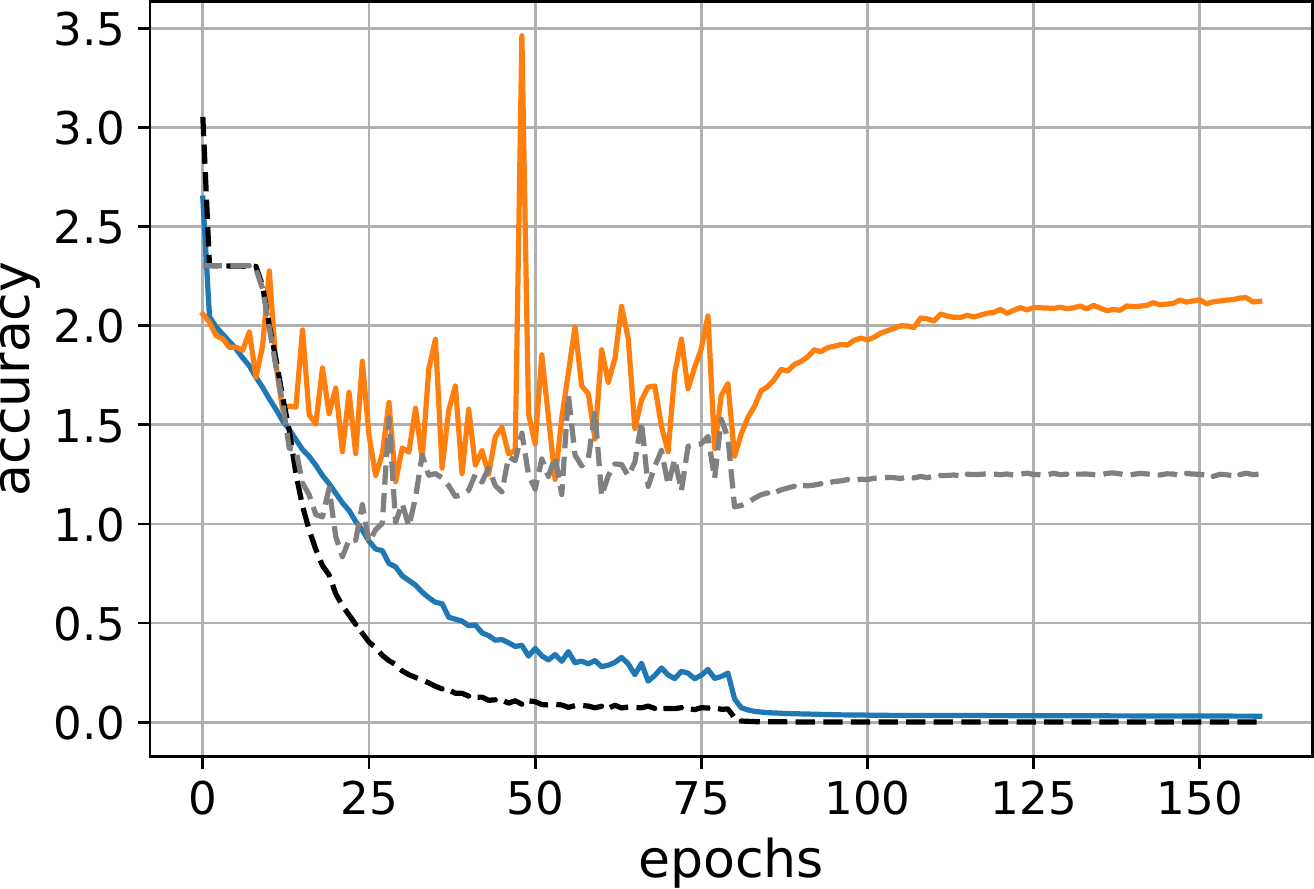}
        \caption[]{Loss ($\sigma$=$32$)}    
        \label{fig:Impl_Symg}
    \end{subfigure}
    \begin{subfigure}[b]{0.245\textwidth}   
        \centering 
        \includegraphics[width=0.95\textwidth]{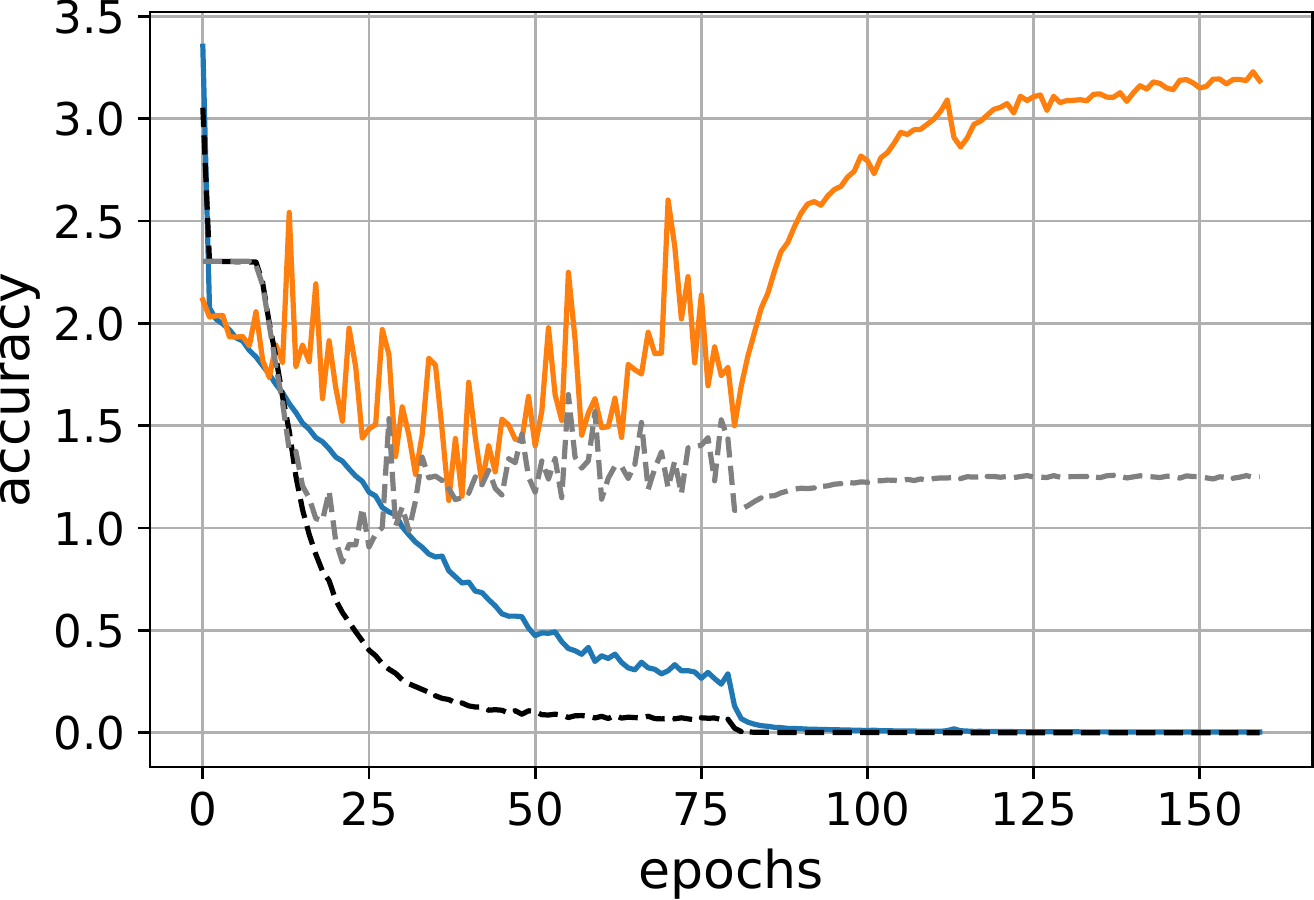}
        \caption[]{Loss ($\sigma$=$64$)}    
        \label{fig:Impl_Symh}
    \end{subfigure}
    \caption[]
    {\small Training curves of the introduced layer for the training (blue) and evaluation (orange) sets of C10~\cite{krizhevsky2009learning}, as obtained by using ResNet-18 architecture~\cite{he2016deep}.
    Black and gray dashed lines depict the corresponding curves of the vanilla FC layer for the same sets.} 
    \label{fig:Impl_Sym}
\end{figure*}

The current section focuses on the implementation of the proposed symmetry in the last layer of a CNN.
The code has been developed using PyTorch 1.4~\cite{paszke2019pytorch}, supporting GPU-enabled operations on an NVIDIA GeForce GTX 1060, 6GB.
According to Section~\ref{sub:PropSym}, regardless the number of classes and the feature space dimensionality, the whole symmetry is placed on a 2-D plane on which all the weights of the last layer lie.
Hence, since the definition of such a plane requires two orthogonal vectors, the trainable parameters of the layer are reduced to the vectors $\bar{v}_1,\bar{v}_2\in\mathcal{F}_d$.
The above two vectors are free in terms of orientation and scale.
In an effort to develop a layer compatible with the supported operations in DL frameworks, such as PyTorch, we follow the Gram-Schmidt orthogonalization~\cite{schmidt1908theorie}
Hence, we calculate the orthonormal vectors $\hat{n}_1,\hat{n}_2\in\mathcal{F}_d$ that form the basis of the rotation plane.
More specifically, $\hat{n}_1$ is the l$_2$-normalized vector of $\bar{v}_1$ and $\hat{n}_2$ is orthonormal to $\hat{n}_1$. 
Consequently, the proposed symmetrical layout is produced by rotating the initial vector $\hat{n}_1$ by $2\pi i/n$, $\forall i\in\mathbb{N}^{<n}$.
The above rotations are conducted in parallel, shaping a GPU-enabled operation.
The code regarding the implementation of the layer is provided in Appendix~\ref{code}, while the whole pipeline for training a CNN is available online\footnote{\textit{https://github.com/IoannisKansizoglou/Symmetrical-Feature-Space}}.

Similarly to the most methods in the field of neural-based feature learning, the introduced layer includes the scaler parameter $\sigma$, which refers to the radius of the hyper-sphere in $\mathcal{F}_d$.
This parameter can be either predefined or it can be learned during the training procedure of the DNN~\cite{ranjan2017l2}.

\section{Experimental Study}\label{ExpAn}

In this section, we proceed with the application of the proposed layer in the broadly known challenge of image classification, exploiting the benchmark database CIFAR-10 (C10)~\cite{krizhevsky2009learning}.
Consequently, an empirical study is demonstrated to highlight the stability inconsistencies observed in the field's state-of-the-art methods, \textit{viz.} \textit{SphereFace}~\cite{liu2017sphereface} and \textit{ArcFace}~\cite{deng2019arcface}.

\subsection{Symmetrical layout convergence} \label{sub:SymLayCon}

\begin{table}
\centering
\caption{Best accuracy percentage (in \%) of each method succeeded in the evaluation set of C10.}
\label{table:EvalAcc}
\resizebox{0.7\linewidth}{!}{%
\renewcommand{\arraystretch}{1.2}
\begin{tabular}{c|c|c}
 Method & $\sigma$ & Accuracy (\%) \\
 \hline
 FC layer & - & $79.77$ \\
 \textit{SphereFace}~\cite{liu2017sphereface} & - & $79.34$ \\
 \textit{ArcFace}~\cite{deng2019arcface} & $64$ & $73.71$ \\
 \hline
 \textbf{Ours} & $8$ & $81.53$ \\
 \textbf{Ours} & $16$ & $82.37$ \\
 \textbf{Ours} & $32$ & $82.22$ \\
 \textbf{Ours} & $64$ & $80.74$ \\
 \hline
\end{tabular}}
\end{table}

\begin{figure}
    \centering
    \begin{subfigure}[b]{0.24\textwidth}
        \centering
        \includegraphics[width=0.99\linewidth]{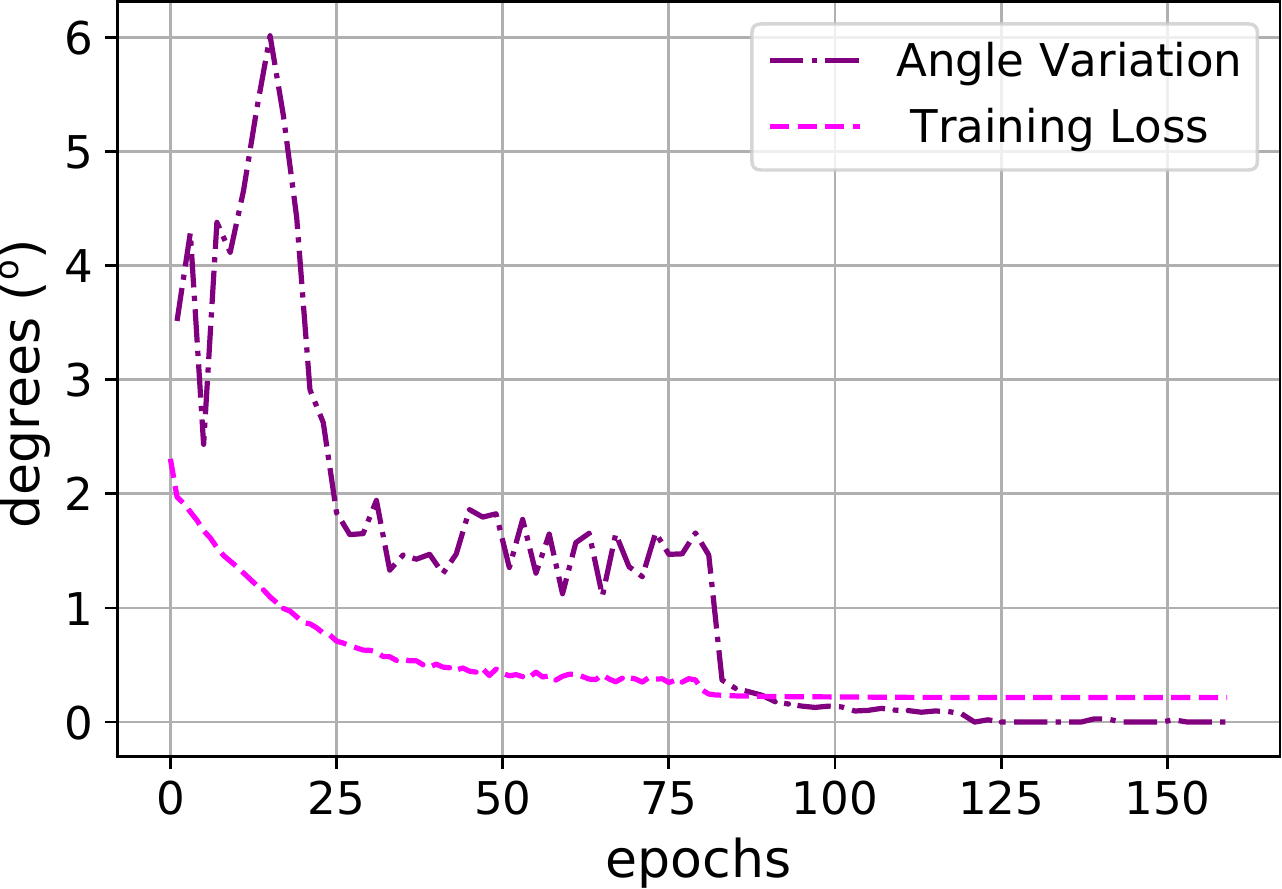}
        \caption{}
        \label{fig:Rotation}
    \end{subfigure}%
	~
    \begin{subfigure}[b]{0.24\textwidth}
        \centering
        \includegraphics[width=0.99\linewidth]{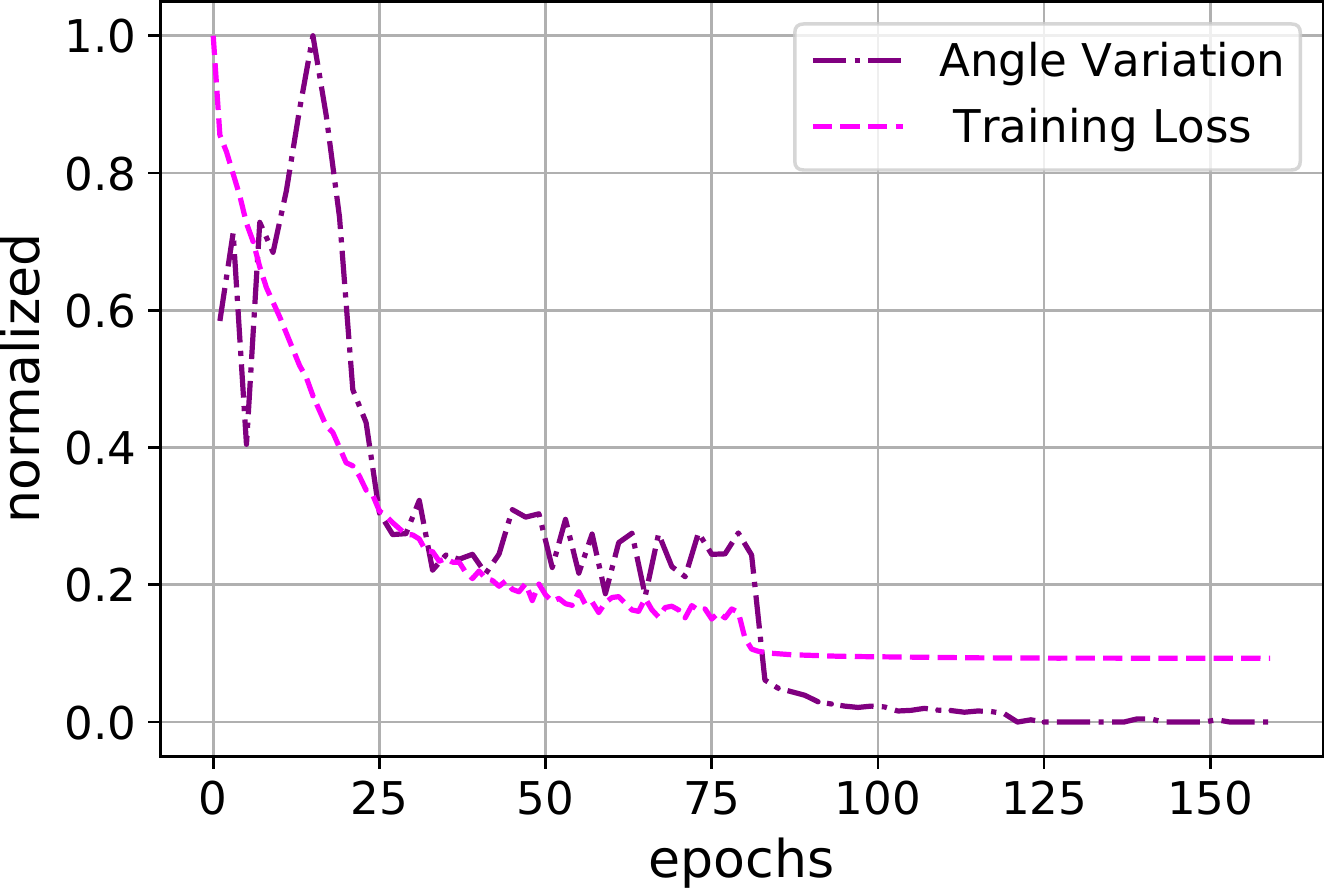}
        \caption{}
        \label{fig:Rotation_n}
    \end{subfigure}
    \caption{\small With purple, the variation of the plane's ($P$) angle (in degrees $^{\circ}$) is illustrated during training the symmetrical layer on C10, while the corresponding training loss is depicted with magenta.
    In (a) both curves are presented in their actual scale, while in (b) they are normalized to further demonstrate their correlation.}
    \label{fig:Rotations}
\end{figure}

In order to evaluate the convergence of the layer, we utilize the widely known ResNet-18 architecture~\cite{he2016deep} and conduct several experiments for different values of $\sigma$ on C10.
To ensure fairness, the training procedure of all experiments lasts 160 epochs with a batch size of $256$.
The Stochastic Gradient Descent (SGD) optimizer is employed, using momentum $0.9$, weight decay $10^{-4}$ and initial learning rate at $0.1$ that decays by an order of magnitude at 50\% and 75\% of the total duration.

In Fig.~\ref{fig:Impl_Sym}, the training curves of the introduced layer on C10 for $\sigma=8,16,32$ and $64$ are demonstrated.
In addition, the corresponding curves for the common FC layer are included to visualize the differences between the two approaches.
Note that the experiments with the FC layer are also conducted following the same training setup.
At first, we highlight the competitive performance of our layer compared against the FC one for all the investigated values of $\sigma$.
In the cases of $\sigma=16$ and $32$, the succeeded evaluation accuracy exceeds by $\approx2.5\%$ the benchmark one, as shown in Table~\ref{table:EvalAcc}.
However, note that the main argument of the current work does not focus on the enhancement of the classification accuracy, rather than on showing that the proposed method sustains state-of-the-art performance. 
Furthermore, by paying attention to the training curves in Fig.~\ref{fig:Impl_Sym}, we understand that the symmetrical layer converges with a slower rate.
The above fact is highly anticipated, due to the restrictions inserted to the trainable parameters to ensure the desired layout.

In Fig.~\ref{fig:Rotations}, the characteristic vector of the plane of symmetry $P$ is monitored.
More specifically, in Fig.~\ref{fig:Rotation} we present the angle between two successive snapshots of the above vector during the training procedure, thus summarizing the variation of its orientation in degrees ($^\circ$).
The curve of the corresponding training loss is also included.
For a more representative comparison, Fig.~\ref{fig:Rotation_n} demonstrates the above curves after normalization in $[0,1]$.

\subsection{Stability issues and comparative study}

The aim of this section is to demonstrate in practice the issues introduced by the common hypothesis $\mathcal{H}$, which is adopted both in \textit{SphereFace}~\cite{liu2017sphereface} and \textit{ArcFace}~\cite{deng2019arcface}, as stated in Section~\ref{RelWor}.
Hence, we proceed with a grid search methodology for the scaler parameter of \textit{ArcFace} with $\sigma$ in $\{4,8,16,32,64\}$, which is adopted by the proposed method, as well.
\textit{SphereFace} is trained only for $\sigma=1$, based on the available implementation of the layer in PyTorch.
Each experiment is conducted three times under exactly the same setup of parameters.
The training setup follows the one proposed in Section~\ref{sub:SymLayCon}.

\begin{figure}
    \centering
    \begin{subfigure}[b]{0.24\textwidth}
        \centering
        \includegraphics[width=0.99\linewidth]{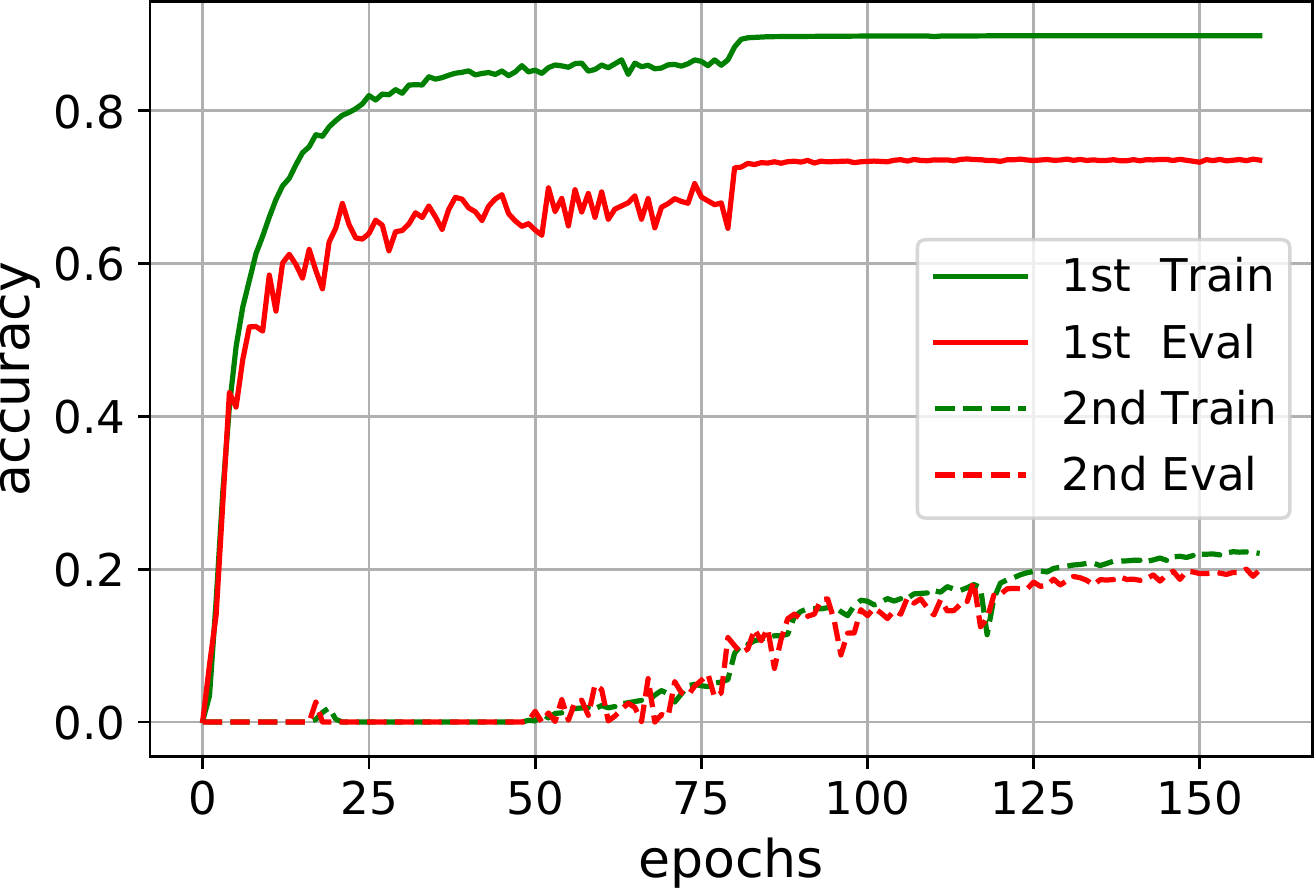}
        \caption{Softmax Accuracy}
        \label{fig:Impl_Arc64A}
    \end{subfigure}%
	~
    \begin{subfigure}[b]{0.24\textwidth}
        \centering
        \includegraphics[width=0.99\linewidth]{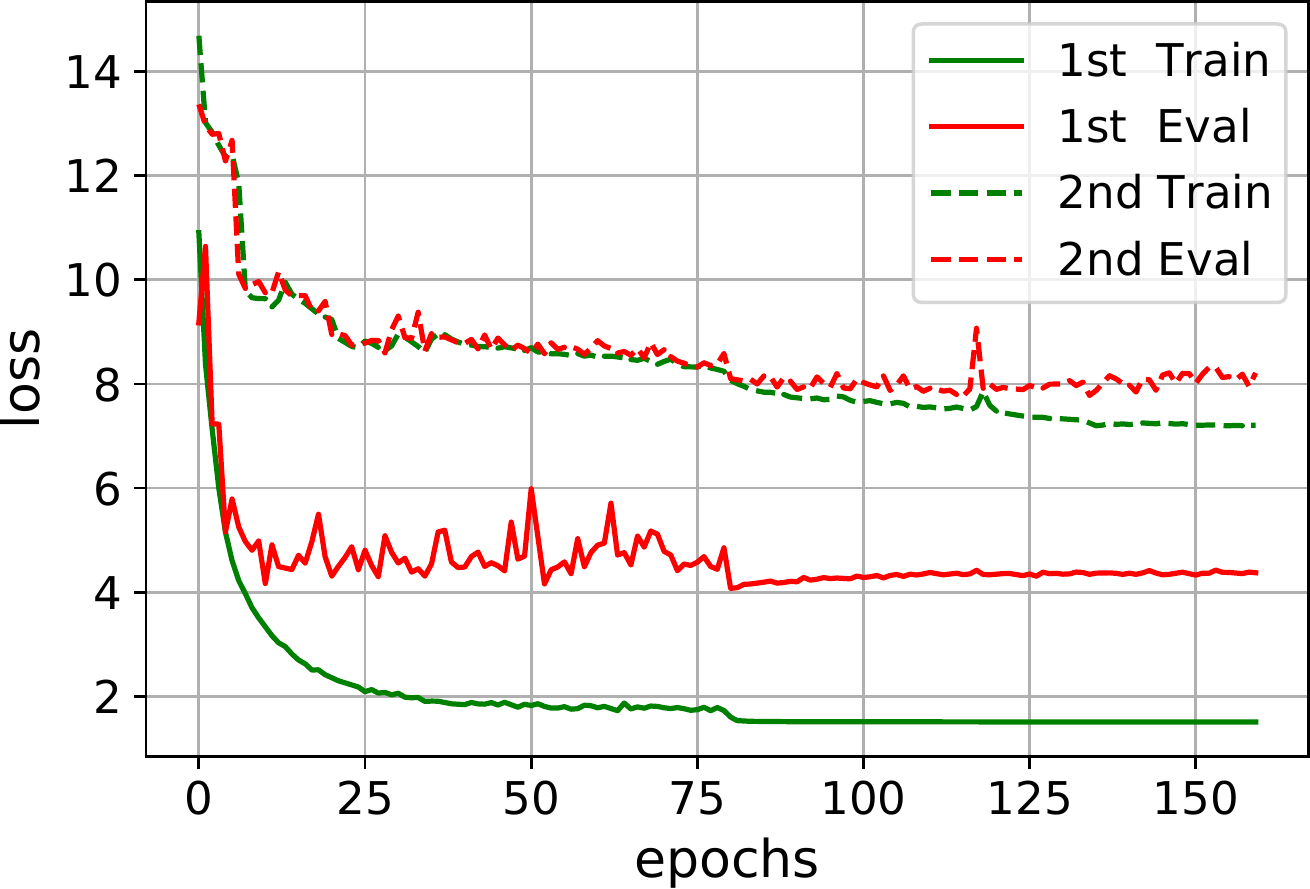}
        \caption{Cross-entropy Loss}
        \label{fig:Impl_Arc64B}
    \end{subfigure}
    \caption{\small Training curves of \textit{ArcFace}~\cite{deng2019arcface} ($\sigma$=$64$, $m$=$0.1$) for the training (green) and evaluation (red) sets of C10, as obtained by using \mbox{ResNet-18}.}
    \label{fig:Impl_Arc64}
\end{figure}

\begin{table}
\centering
\caption{Training results under different experimental setups of \textit{ArcFace}~\cite{deng2019arcface}, as obtained by using ResNet-18 on C10.}
\label{table:StabIsArc}
\resizebox{0.85\linewidth}{!}{%
\renewcommand{\arraystretch}{1.2}
\begin{tabular}{c|c|c|c|c} 
 $\sigma$ & $m$ & $Acc_1$ ($\%$) & $Acc_2$ ($\%$) & $Acc_3$ ($\%$) \\
 \hline
 $4$ & $0.1$ & x & x & x \\
 $8$ & $0.1$ & x & x & x \\
 $16$ & $0.1$ & x & x & x \\
 $32$ & $0.1$ & x & $10.00$ & x \\
 $64$ & $0.1$ & $20.02$ & $73.71$ & x \\
\end{tabular}}
\end{table}

The obtained results for \textit{ArcFace} are depicted in Table~\ref{table:StabIsArc}.
The notation x is employed to represent the cases, where the training loss diverges from finite values.
In any other case, the best evaluation accuracy (in $\%$) is kept.
By examining Table~\ref{table:StabIsArc}, the reader can discover the unstable and simultaneously different behavior of such a method even under identical training and hyper-parameters setup.
We argue that the above fact is due to the error inherited by $\mathcal{H}$.
The above renders in turn the investigated methods highly dependent upon weights initialization since the higher the asymmetry of the initial weight vectors, the higher the divergence between the orientation of a weight and the actual centre of the corresponding class.

Paying more attention to the particularly interesting case of $\sigma=64$, we observe that for two distinct repetitions of the same experiment, the obtained evaluation accuracy ends up to different finite values.
Hence, we illustrate in Fig.~\ref{fig:Impl_Arc64} the training curves of those two repetitions, demonstrating the method's susceptibility to weights initialization.
As a typical instance, the reader can easily identify in the second case, depicted with dashed lines, the large delay until the onset of the convergence.

\begin{figure}
    \centering
    \begin{subfigure}[b]{0.24\textwidth}
        \centering
        \includegraphics[width=0.99\linewidth]{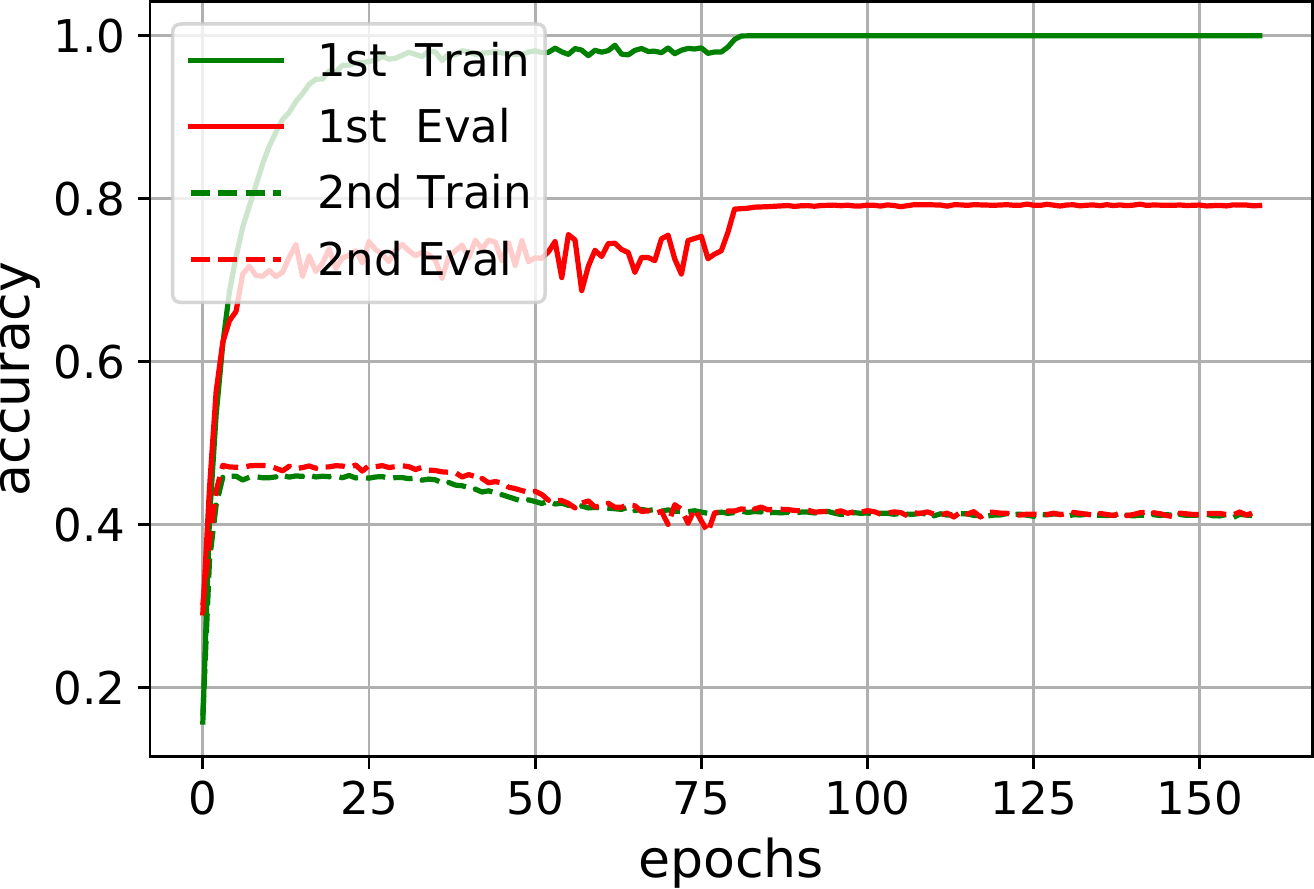}
        \caption{Softmax Accuracy}
        \label{fig:Impl_Sph1A}
    \end{subfigure}%
	~
    \begin{subfigure}[b]{0.24\textwidth}
        \centering
        \includegraphics[width=0.99\linewidth]{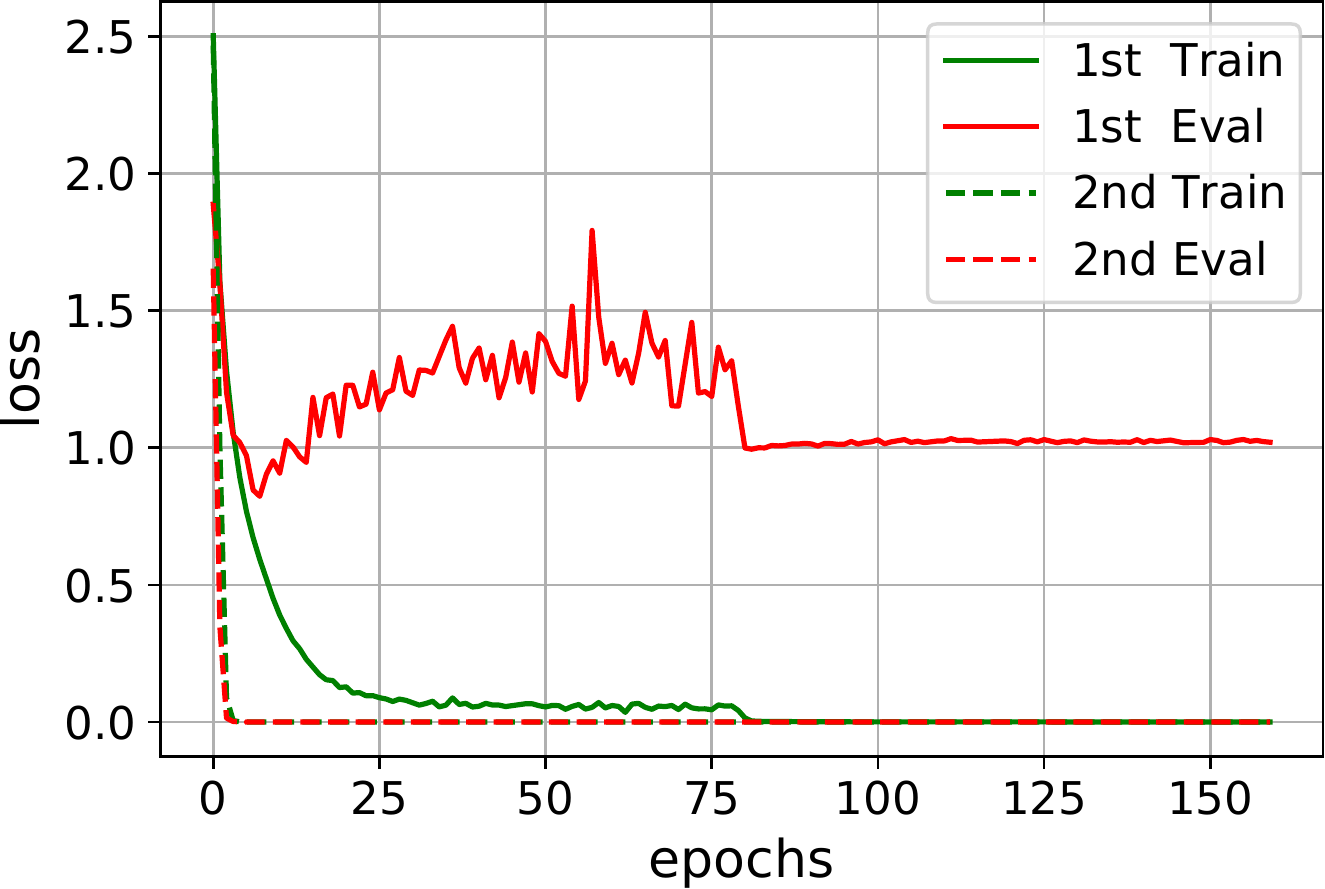}
        \caption{Cross-entropy Loss}
        \label{fig:Impl_Sph1B}
    \end{subfigure}
    \caption{\small Training curves of \textit{SphereFace}~\cite{liu2017sphereface} ($\sigma$=$1$) for the training (green) and evaluation (red) sets of C10, as obtained by using \mbox{ResNet-18}.}
    \label{fig:Impl_Sph1}
\end{figure}

\begin{table}
\centering
\caption{Training results of \textit{SphereFace}~\cite{liu2017sphereface}, as obtained by using ResNet-18 on C10.}
\label{table:StabIsSph}
\resizebox{0.7\linewidth}{!}{%
\renewcommand{\arraystretch}{1.2}
\begin{tabular}{c|c|c|c} 
 $\sigma$ & $Acc_1$ ($\%$) & $Acc_2$ ($\%$) & $Acc_3$ ($\%$) \\
 \hline
 $1$ & $47.31$ & $79.34$ & x \\
\end{tabular}}
\end{table}

Eventually, the same experimental procedure is employed for the case of \textit{SphereFace} using $\sigma=1$.
Again, the obtained results are displayed in Table~\ref{table:StabIsSph}  and differ under different repetitions of the same training setup, empirically confirming our statement.
Then, we further display in Fig.~\ref{fig:Impl_Sph1} the training curves of the two repetitions that end up to finite values of evaluation accuracy.
Here, we observe that both training and evaluation accuracy display a common plateau since they reach a specific value, which cannot be exceeded by the CNN.

\subsection{Time Complexity}

As a final assessment, we focus on time efficiency, ascertaining that the proposed implementation adds no particular complexity as compared against a simple Fully Connected (FC) solution.
In Table~\ref{table:TimDur}, the mean and standard deviation values of the corresponding seconds per epoch are computed among three repetitions of the same experiment, using an FC layer, \textit{SphereFace}, \textit{ArcFace} and ours.
As shown, all methods require about the same training time, ensuring that the proposed implementation is suitable for large-scale applications.

\begin{table}
\centering
\caption{Training duration (in $sec/epoch$) of the introduced layer compared against \textit{ArcFace}, \textit{SphereFace} and the common FC layer.}
\label{table:TimDur}
\resizebox{0.75\linewidth}{!}{%
\renewcommand{\arraystretch}{1.2}
\begin{tabular}{c|c}
 Method & Duration ($sec/epoch$) \\
 \hline
 FC layer & $79.79\pm0.15$ \\
 \textit{SphereFace}~\cite{liu2017sphereface} & $81.66\pm2.39$ \\
 \textit{ArcFace}~\cite{deng2019arcface} & $81.62\pm3.75$ \\
 \textbf{Ours} & $80.14\pm0.08$ \\
 \hline
\end{tabular}}
\end{table}

\section{Conclusion} \label{Conc}

In drawing to a close, the paper at hand deals with the empirical assumption adopted by the vast majority of recent methods in the field of neural-based feature learning.
The aforementioned hypothesis ($\mathcal{H}$) claims the coincidence between the last FC layer's weight vector with the geometrical centre of the corresponding class.
Following a theoretical approach, we prove the refutability of $\mathcal{H}$, given a random distribution of the weight vectors in the feature space $\mathcal{F}_d$, which refers to the input space of the output layer.
Consequently, a specific symmetrical layout for the weight vectors is proposed that is proved to satisfy $\mathcal{H}$ for any dimensionality $d$ of space $\mathcal{F}_d$.

Proceeding with our empirical study, the implementation of the proposed symmetry is described, which can be easily adopted as a custom layer in widespread deep learning frameworks, like PyTorch.
The code of the layer is openly provided in Appendix~\ref{code}.
Then, several experiments are conducted to demonstrate the convergence capabilities of the layer.
We further demonstrate that the required training duration is similar to the one of a common FC layer.
At competitive and occasionally beneficiary levels are also the achieved accuracy values of the proposed layer compared against the FC one, as well as two of the most widespread models in the field, \textit{viz.}, \textit{SphereFace} and \textit{ArcFace}.
Finally, the impact of the false supposition in stability issues is empirically demonstrated within the above methods in the field, by evaluating their achieved evaluation accuracy between multiple repetitions of the same training setup .

As part of future work, we aim to exploit the formulation of the proposed symmetrical layout in a feature learning task, such as the challenge of face verification.
In specific, the combination of the proposed layer with the rules of state-of-the-art schemes, such as \textit{ArcFace}, can be investigated, as an attempt to enhance the feature learning capacity of a DNN.



\ifCLASSOPTIONcaptionsoff
  \newpage
\fi


\bibliographystyle{IEEEtran}
\bibliography{IEEEabrv,root}

\begin{IEEEbiography}[{\includegraphics[width=1in,height=1.25in,clip,keepaspectratio]{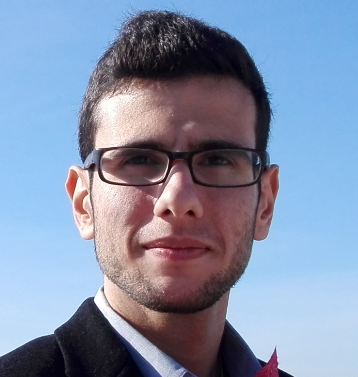}}]{Ioannis Kansizoglou}
received the diploma degree from the Department of Electrical and Computer Engineering, Aristotle University of Thessaloniki, Greece, in 2017.
Currently, he is working toward the PhD degree in the Laboratory of Robotics and Automation, Department of Production and Management Engineering, Democritus University of Thrace, Greece, working on emotion analysis and its application in robotics.
\end{IEEEbiography}

\begin{IEEEbiography}[{\includegraphics[width=1in,height=1.25in,clip,keepaspectratio]{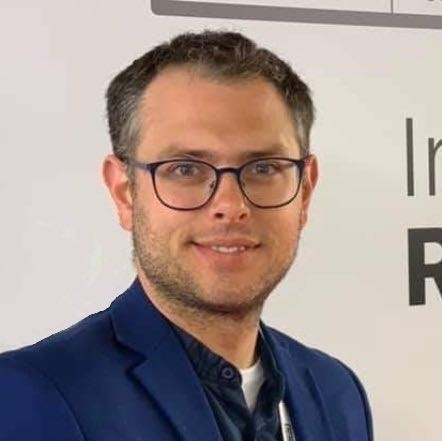}}]{Loukas Bampis}
received the diploma degree in Electrical and Computer Engineering and PhD degree in machine vision and embedded systems from the Democritus University of Thrace (DUTh), Greece, in 2013 and 2019, respectively.
He is currently a postdoctoral fellow in the Laboratory of Robotics and Automation (LRA), Department of Production and Management Engineering, DUTh.
His work has been supported through several research projects funded by the European Space Agency, the European Commission and the Greek government.
His research interests include real-time localization and place recognition techniques using hardware accelerators and parallel processing.
\end{IEEEbiography}

\begin{IEEEbiography}[{\includegraphics[width=1in,height=1.25in,clip,keepaspectratio]{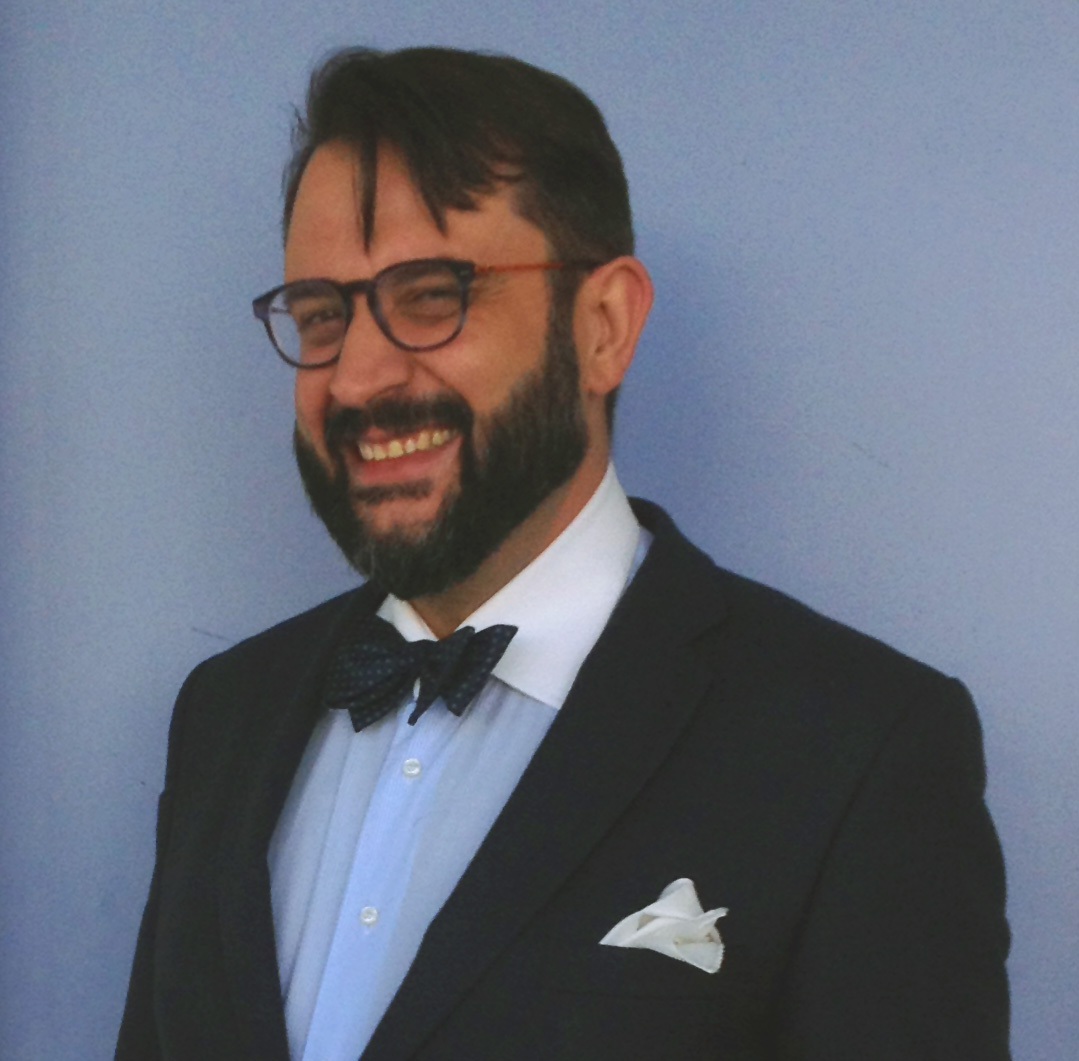}}]{Antonios Gasteratos}
received the MEng and PhD degrees from the Department of Electrical and Computer Engineering, Democritus University of Thrace (DUTh), Greece.
He is a professor and head of Department of Production and Management Engineering, DUTh, Greece. He is also the director of the Laboratory of Robotics and Automation (LRA), DUTh and teaches the courses of robotics, automatic control systems, electronics, mechatronics and computer vision.
During 1999–2000 he was a visiting researcher at the Laboratory of Integrated Advanced Robotics (LIRALab), DIST, University of Genoa, Italy.
He has served as a reviewer for numerous scientific journals and international conferences. He is a subject editor at Electronics Letters and an associate editor at the International Journal of Optomecatronics and he has organized/coorganized several international conferences.
His research interests include mechatronics and robot vision. He has published more than 220 papers in books, journals and conferences.
He is a fellow of IET and a senior member of the IEEE.
\end{IEEEbiography}

\vfill

\appendices
\section{}\label{code}

\begin{lstlisting}
# PyTorch code 

import torch
import torch.nn as nn
import torch.nn.functional as F

class SymmetricalLayer(nn.Module):
    def __init__(self, input_features,
    						num_classes, scaler, device):
        super(SymmetricalLayer, self).__init__()
        
        self.weight = nn.Parameter(
        				torch.FloatTensor(2, input_features)
        			)
        nn.init.uniform_(self.weight)
        thetas = torch.arange(num_classes, dtype=torch.float32)
        self.thetas = (2*math.pi*thetas / thetas.shape[0]).to(device)
        self.I = torch.eye(input_features).to(device)
        self.input_features = input_features
        self.num_classes = num_classes
        self.s = scaler
    
    def rotateNd(self, v1, v2):

        n1 = v1 / torch.norm(v1)
        v2 = v2 - torch.dot(n1,v2) * n1
        n2 = v2 / torch.norm(v2)
        
        ger_sub = torch.ger(n2,n1) - torch.ger(n1,n2)
        ger_add = torch.ger(n1,n1) + torch.ger(n2,n2)
        sin_th = torch.unsqueeze(
        			torch.unsqueeze(
        				torch.sin(self.thetas),dim=-1),
        		dim=-1)
        cos_th = torch.unsqueeze(
        			torch.unsqueeze(
        				torch.cos(self.thetas)-1,dim=-1),
        		dim=-1)
        R = self.I + ger_sub*sin_th + ger_add*cos_th
        
        return torch.einsum('bij,j->bi',R,n1)        
        
    def forward(self, input_x):
        x = F.normalize(input_x)
        W = F.normalize(self.weight)
        Ws = self.rotateNd(W[0],W[1])
        cosine = F.linear(x, Ws)
        output = self.s*cosine
        
        return output

\end{lstlisting}




\end{document}